\documentclass[journal]{IEEEtran}
\usepackage{graphics,graphicx}
\usepackage{amsmath,amssymb,amsthm}
\usepackage{multirow}
\usepackage{cases}
\usepackage{array}
\usepackage{algorithm,algorithmic}
\usepackage{color,colortbl,xcolor}
\usepackage[graphicx]{realboxes}
\usepackage{booktabs,url}
\usepackage[tight,footnotesize,sf,SF]{subfigure}
\usepackage{bigstrut}

\newcommand{\fref}[1]{Fig. \ref{#1}}
\newcommand{\sref}[1]{Section \ref{#1}}
\newcommand{\tref}[1]{TABLE \ref{#1}}
\newcommand{\eref}[1]{Eq. (\ref{#1})}

\begin{document}
\title{Large-scale matrix optimization based multi microgrid topology design with a constrained differential evolution algorithm}

\author{
Wenhua Li,
Shengjun Huang,
Tao Zhang,
Rui Wang, \emph{Senior Member, IEEE},
and Ling Wang

\thanks{This work was supported by the National Science Fund for Outstanding Young Scholars (62122093), the National Natural Science Foundation of China (72071205), the Scientific Key Research Project of the National University of Defense Technology (ZZKY-ZX-11-04) and the Ji-Hua Laboratory Scientific Project (X210101UZ210).}
\thanks{Wenhua Li is with the College of Systems Engineering, National University of Defense Technology, Changsha, China, 410073, e-mail: liwenhua@nudt.edu.cn.}
\thanks{Shengjun Huang, Tao Zhang and Rui Wang are with the College of Systems Engineering, National University of Defense Technology, Changsha, China, and Hunan Key Laboratory of Multi-energy System Intelligent Interconnection Technology, Changsha, 410073, China}
\thanks{Ling Wang is with the Department of Automation, Tsinghua University, Beijing, 100084, P.R. China. }
\thanks{Corresponding author: Rui Wang (email: ruiwangnudt@gamil.com). }
}
\maketitle

\begin{abstract}
Binary matrix optimization commonly arise in the real world, e.g., multi-microgrid network structure design problem (MGNSDP), which is to minimize the total length of the power supply line under certain constraints. Finding the global optimal solution for these problems faces a great challenge since such problems could be large-scale, sparse and multimodal. Traditional linear programming is time-consuming and cannot solve nonlinear problems. To address this issue, a novel improved feasibility rule based differential evolution algorithm, termed LBMDE, is proposed. To be specific, a general heuristic solution initialization method is first proposed to generate high-quality solutions. Then, a binary-matrix-based DE operator is introduced to produce offspring. To deal with the constraints, we proposed an improved feasibility rule based environmental selection strategy. The performance and searching behaviors of LBMDE are examined by a set of benchmark problems.
\end{abstract}


\IEEEpeerreviewmaketitle

\section{Introduction}
\label{sec_introduction}

With the increasing attention on renewable energy, microgrid technology \cite{li2021sizing,li2019multi} has been successfully implemented in many aspects, e.g., manufacturing factories, farms \cite{8239827} and industrial parks \cite{9504619}. For some remote areas which can not access the power grid, stand-alone microgrids \cite{8171865} have been an effective tool to maintain the stable running of the power system. However, since there is unavoidable randomness in renewable energy generation, e.g., extreme weather and power supply device failure, the continuous and stable running face big challenges, especially for some important facilities \cite{9637999}. 

Therefore, building a multi-microgrid network is necessary to improve the robustness of the system. To be specific, several microgrids can be seen as nodes, which run independently in normal times \cite{sym14051014}. Then, once the power generation of one certain microgrid is broken or unable to supply the energy consumption, the pre-set power supply circuits will be activated/turn-on to maintain the system running. Therefore, designing the microgrid power supply circuits network becomes an important issue, known as multi-microgrid network structure design problems (MNSDPs), which aims to minimize the total length of power supply circuits under certain constraints \cite{camacho2019optimal}.

Generally speaking, a solution of an MNSDP can be presented by a binary matrix. Such problems usually exist in real-world engineering problems, such as topology design \cite{wang2005structural}, power supply network reconfiguration \cite{li2021sizing} and logistics optimization \cite{8950401}, which is known as binary matrix optimization problems (BMOPs) \cite{zhang2018deterministic,curtis2006determinant,kumar2019faster}. Without loss of generality, a BMOP can be expressed as:

\begin{equation}
\begin{gathered}
\text{Minimize}~~f(\mathbf{x}) \\
s.t.~~~~ g(\mathbf{x}) \leq 0, \\
~~~~~~~~ h(\mathbf{x}) = 0, \\
\mathbf{x}_{i,j} \in \{0,1\}, i \in [1,m], j \in [1,n]
\end{gathered}
\label{eq_MOP}
\end{equation}
where $\mathbf{x}$ denotes the binary decision variables and $\mathbf{x}$ is a decision vector that consists of $m*n$ decision variables $x_i$. A solution $\mathbf{x_a}$ is considered as a global optimal solution, $iff$ $\forall b=1,2,...,N, f(\mathbf{x_a}) \leq f(\mathbf{x_b})$, where solutions $\mathbf{x_a}$ and $f(\mathbf{x_b})$ are feasible. 

Over the last several decades, evolutionary algorithms (EAs) have been successfully utilized to solve many complex real-world engineering problems \cite{li2019knee,YAO2022108145}. However, most of the primitive EAs mainly focus on continuous optimization problems. Then, many approaches have been proposed to solve discrete and binary optimization problems based on the state-of-the-art EAs, e.g., artificial bee colony algorithm \cite{kiran2015continuous}, grey wolf optimization \cite{al2019binary}, particle swarm optimization \cite{khanesar2007novel} and Jaya algorithm \cite{aslan2019jayax}. In general, these approaches deal with binary decision variables in the following two ways: using real values in $[0,1]$ to encode the solution and transform them into 0 or 1; utilizing novel operators to generate solutions, e.g., $xor$ operator \cite{aslan2019jayax}.

The above-mentioned algorithms are effective and efficient in dealing with low-dimension problems which consist only of 2-30 decision variables. However, as the studied problems become more and more complex, the number of decision variables could be huge, which is known as large-scale optimization problems \cite{hager2013large,jian2021region,li2021adaptive}. So far, utilizing EAs to solve such large-scale problems is challenging since the decision space is huge and operators such as simulated binary crossover (SBX) and polynomial mutation (PM) are low-efficient in exploring the whole decision space. Therefore, much research has been carried out to deal with large-scale binary optimization problems. Kong et al. \cite{kong2015simplified,kong2015solving} proposed the binary harmony search algorithm (SBHS) to solve large-scale knapsack problems with up to 10,000 items by changing the process of improvisation. In addition, a new self-adaptive binary variant of a differential evolution algorithm \cite{BANITALEBI2016487} is proposed to solve large-scale learning-based problems, which uses an adaptive mechanism to select ways to generate solutions. In \cite{panwar2018binary}, the binary grey wolf optimizer is adopted to solve large-scale unit commitment problems, in which a novel crossover operation and continuous estimation approach are proposed to enhance the algorithm performance. 

For now, the efficiency and effectiveness of the existing large-scale EAs in solving BMOPs have not been well studied. Therefore, to better address the large-scale BMOPs, we proposed a novel differential evolution algorithm based on improved feasible rules. Generally speaking, the main contributions of this study can be concluded as follows:
\begin{itemize}
\item The mathematical model of a multi-microgrid network structure design problem is established, which considers three different types of nodes. In addition, test instances can be easily generated by adjusting user-defined parameters. The decision variable of this problem is a symmetry binary matrix.
\item To accelerate the convergence of EAs in dealing with large-scale BMOPs, a binary-matrix-based DE operator is proposed to generate solutions, which can be easily embedded into the existing EAs.
\item To better balance the constraints and objective values during the evolution, an improved feasible rule based strategy is proposed, which can well maintain the diversity of solutions while exerting pressure on the evolution.
\item Based on the proposed operator and improved feasible rule based strategy, a novel EA is proposed, termed LBMDE. Compared to the traditional mixed integer programming method (MIP) and other state-of-the-art EAs, experimental results show that LBMDE is competitive in dealing with large-scale BMOPs.  
\end{itemize}

The remainder of this work is structured as followed: a brief review of the network structure design problems and the evolutionary algorithm for BMOPs are made in \sref{sec_relatedwork}, followed by \sref{sec_model} and \sref{sec_algorithm}, which illustrate the proposed mathematical model of the MNSDP and the proposed differential evolution algorithm in detail respectively. Then, we conduct several experiments to examine the effectiveness of our proposed model and algorithm in \sref{sec_exp}. Finally, we concluded this work and raised several possible ways for future research in \sref{sec_conclusion}.

\section{Preliminary works}
\label{sec_relatedwork}

\subsection{Network structure design problems}
\label{sec_nsdp}
\begin{figure}[tbph]
	\begin{center}
		\includegraphics[width=3.5in]{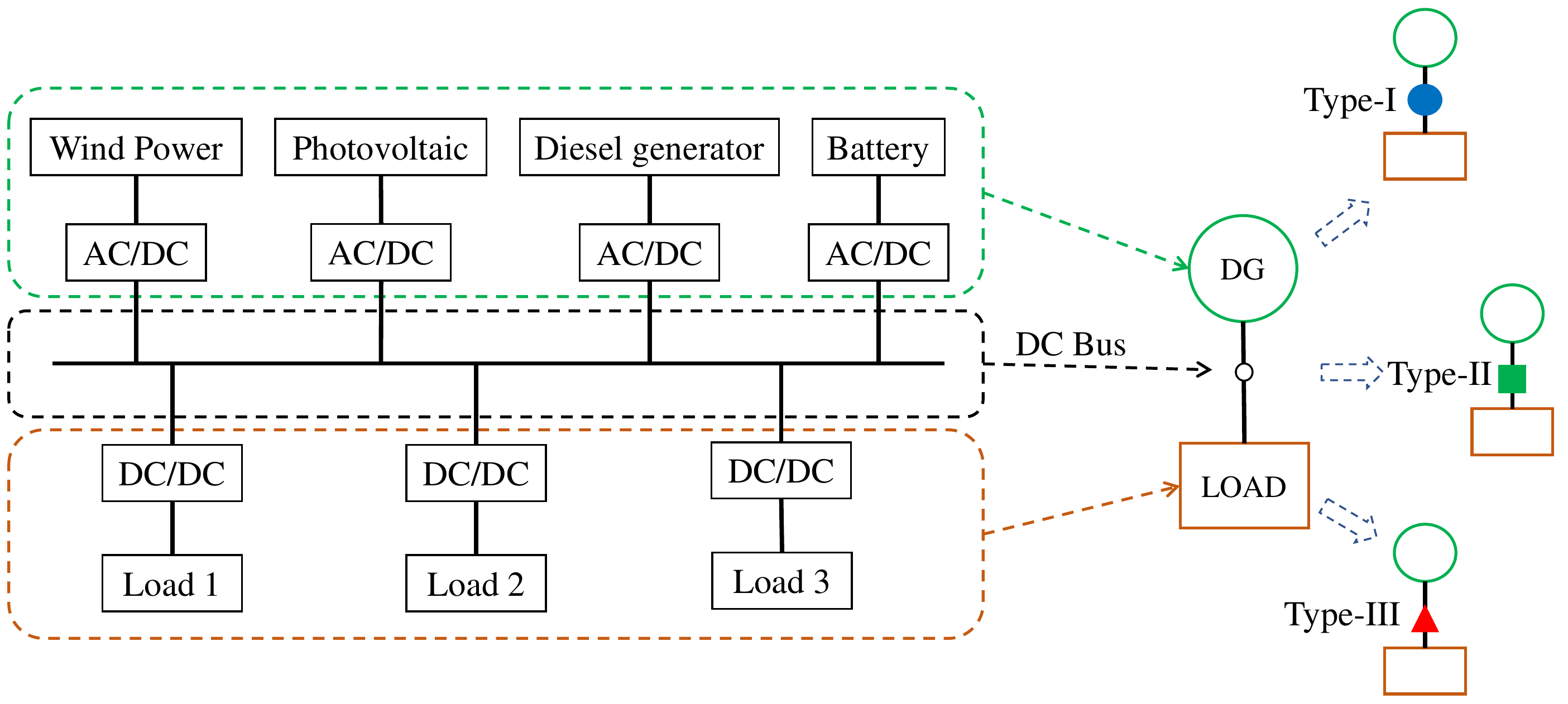}
		\caption{Abstract diagram of nodes in the multi-microgrid network system.}
		\label{fig_problem}
	\end{center}
\end{figure}

As a typical power insurance network, the multi-microgrid network system (MGNS) \cite{wenzhi2022hierarchical,ge2021dynamic} consists of multiple stand-alone running microgrids. For the convenience of presentation and description, in the later of this study, each microgrid will be considered as a node, as shown in \fref{fig_problem}. 

As we can see from \fref{fig_problem}, there are one DC bus, four power sources (wind power, photovoltaic, diesel generator and battery system) and three loads in a stand-alone microgrid system. Generally speaking, the total power generation, known as DG, is larger than the power consumption (loads) in a stand-alone microgrid system to ensure stable running. To better illustrate the problem ins this study, we use green circles, orange boxes, and lines with a black circle to represent the total power generation, total loads and the DC bus, respectively. In addition, the blue circle, green square and red triangle are used to represent the type-I, type-II and type-III nodes respectively, which will be further illustrated in the following parts.

Each node in the MGNS runs alone if there is no malfunction. Once the power generation is broken or insufficient to cover the loads, then the pre-set power supply circuits connected to the failed node will be activated to transfer the excess power from the neighbor nodes \cite{zhou2021multi}. To simplify the process, we assume that once the DG in the node fails, the circuits related to the node are immediately connected to provide energy support.

Generally speaking, we can use failure probability to describe the stability of a system \cite{pourghasem2022new}. However, it is challenging to evaluate the exact failure probability of a single node. Therefore, the $N-K$ method \cite{chen2005identifying,beyza2021integrated} is introduced, which means that a system with $N$ nodes (components) can still work normally after $K$ nodes are damaged. The $N-K$ method measures the system stability from a systematical point of view, which assumes that all nodes have the same failure probability or equal importance. However, this is not always true for real-world MGNS \cite{sym14051014}.

\begin{figure}[tbph]
	\begin{center}
		\includegraphics[width=3.5in]{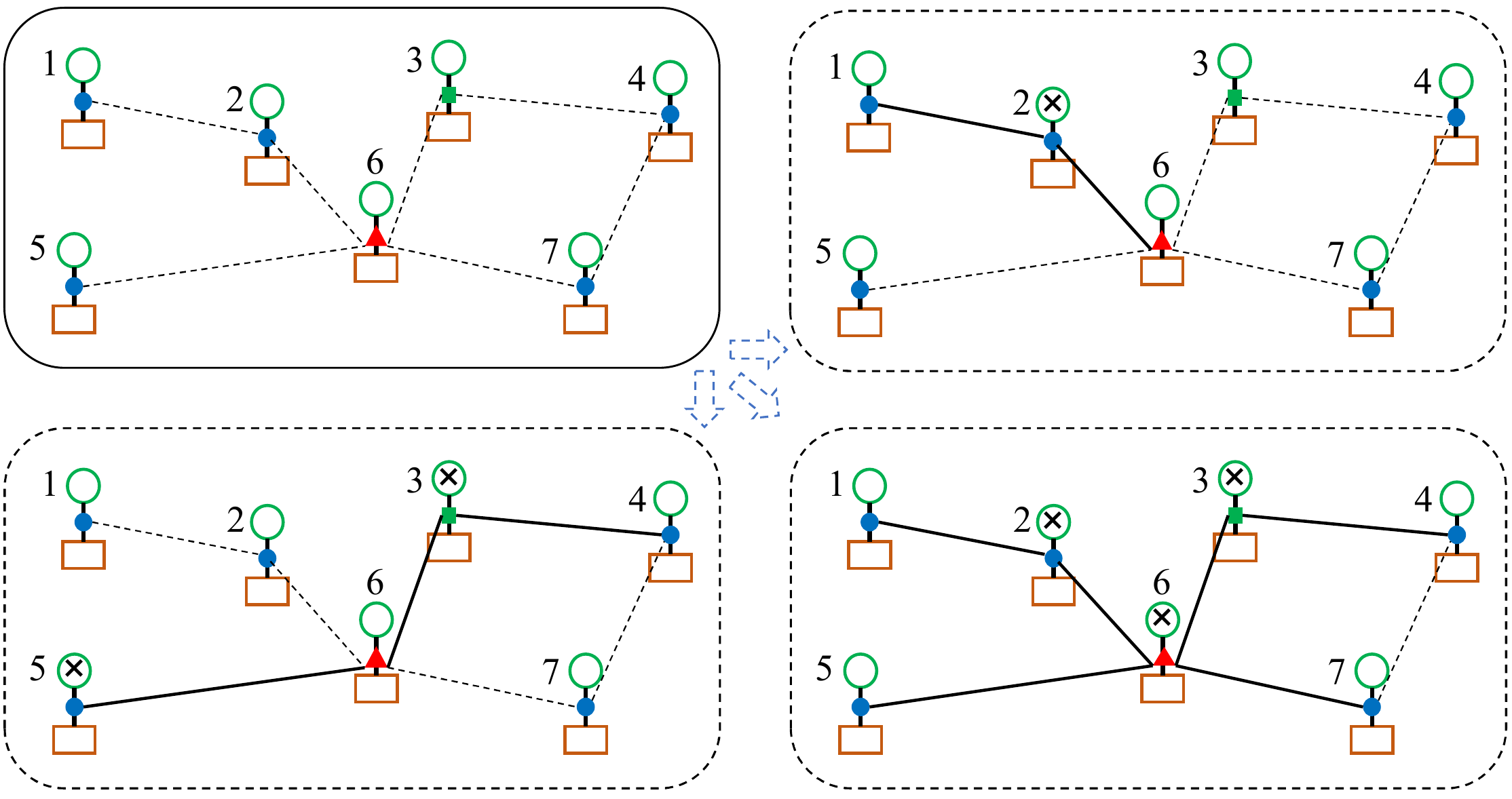}
		\caption{Schematic diagram of topological transformation triggered by microgrid node failure, where the dotted lines are pre-set power supply circuits.}
		\label{fig_errortype}
	\end{center}
\end{figure}

In order to present the differences in node reliability requirements, in this study, each node $i,\ i \in [1,N]$ is assigned with a criterion $N-K_i$ according to its special situation, which means that node $i$ can work normally when $K_i$ nodes are breakdown. According to different $K$, nodes can be categorized as different types. To be specific, in this study, $K$=1, 2 and 3 correspond to type-I, type-II and type-III nodes respectively. The larger $K$ is, the higher importance of the node is \cite{ren2008using}. \fref{fig_errortype} shows several examples of the topological transformation triggered by microgrid node failure in an MGNS. The upper left figure is the normal running state of the MGNS, which consists of five type-I (nodes 1, 2, 4, 5 and 7), one type-II (node 3) and one type-III (node 6) nodes. The upper right figure shows a simple situation that one type-I node (node 2) has a failure. Then, the pre-set circuits will connect node 1 and node 6 to provide energy to node 2. For the situation in the bottom right figure, nodes 2, 3 and 6 have failed simultaneously. As a result, the circuits between these nodes and their neighbors are activated.

According to the above description, the nodes in the MGNS have characteristics that are distributed in normal time and interconnected in failure time. Therefore, in order to keep a MGNS stably runs with a low cost, it is of significant importance to find the optimal network structure, which is known as the multi-microgrid network structure design problems (MNSDPs).

\subsection{Evolutionary algorithms}

EAs have been an effective and efficient tool to solve complex and non-linear problems. However, although many real-world engineering problems belong to matrix optimization problems, few studies have systematically discussed this kind of problem. For reasons, some problems can also be encoded with several layers. For example, in many studies for job-shop scheduling problems \cite{li2019knee,he2021optimising}, the order of the processes and the corresponding machines are separately encoded with a two-layer coding system. In \cite{absil2009optimization}, Absil et al. studied how to exploit the special structure of such problems to develop efficient numerical algorithms. Then, Kang \cite{Kang2013} et al. presented a matrix-based automated concept generation method and used ant colony optimization to optimize the generated conceptual candidate solutions. 

For BMOPs, many studies focus on enhancing the performance of the existing EAs during the last several decades. Banitalebi et al. \cite{BANITALEBI2016487} proposed a self-adaptive binary DE algorithm (SabDE) to solve large-scale binary optimization problems. Tashi et al. \cite{panwar2018binary} proposed a binary version of hybrid grey wolf optimization to solve the feature selection problems, which is demonstrated effective according to the experimental results. In addition, many other EAs have been utilized to develop algorithms for binary optimization problems, e.g., angle modulated DE (AMDE) \cite{1688535}, discrete binary DE (DBDE) \cite{peng2008solving}, binary hybrid topology particle swarm optimization (BHTPSO-QI) \cite{beheshti2015memetic}, binary bee colony optimization (binABC) \cite{jia2014binary}, simplified binary harmony search algorithm (SBHS) \cite{kong2015solving} and Binary Quantum-Inspired Gravitational Search Algorithm (BQIGSA) \cite{nezamabadi2015quantum}. Recently, the Jaya algorithm with $xor$ operator for binary optimization (JayaX) \cite{aslan2019jayax} is proposed and verified by CEC 2015 numeric functions. As we can see, the binary version of EAs has been well studied. Then, in \cite{patle2018matrix}, a matrix-binary codes-based genetic algorithm (MGA) is proposed that uses the binary codes through a matrix for mobile robot navigation in a static and dynamic environment. So far, designing algorithms for large-scale BMOPs is still in its infancy. However, such methods/approaches are imperative.

\section{Mathematical model of the MNSDP}
\label{sec_model}
\subsection{Constraints for type-I nodes}
As we describe in \sref{sec_nsdp}, for type-I nodes, according to the definition of $N-1$, it is required that the current node can still work normally when the system loses 1 node. Therefore, the worst scenario is that DG of the current node is zero, and the neighbor nodes are required to provide sufficient support for the load of the current node. Therefore, for a type-I node $i$, the following constraints should be satisfied.

\begin{equation}
S_{i} \geq L_{i}, \forall i \in V_{I},
\end{equation}
\begin{equation}
S_{i}=\sum_{j=1, j \neq i}^{n}\left(G_{j}-L_{j}\right) x_{ij},
\end{equation}
where $n$ and $V_{I}$ are the total number of nodes and the set of type-I nodes, $S_{i}$ is the energy support that all neighbor nodes of node $i$ can provide; $L_{i}$ and $G_{i}$ are the load and power generation of node $i$ respectively. $x_{ij}$ is the decision variable, where $x_{ij}=1$ means that there is connection between node $i$ and node $j$.

\subsection{Constraints for type-II nodes}
For type-II nodes, it is required that the current node can still work normally when the system loses 2 nodes arbitrarily. Similar to the case of type-I nodes, the worst scenario occurs when the power supply of the current node and its neighbor node is damaged, and loads of the two nodes need to be supported by their neighbor nodes. Therefore, a type-II node $j$ should satisfy the following constraints.

\begin{equation}
S_{j}+S_{i}-r_{i j}-s_{i j} \geq L_{j}+L_{i}, \forall j \in V_{I I}, \forall i \in V, i \neq j
\end{equation}
\begin{equation}
r_{i j}=x_{i j}\left(G_{i}-L_{i}+G_{j}-L_{j}\right)
\end{equation}
\begin{equation}
s_{i j}=\sum_{k=1, k \neq i, k \neq j}^{n} x_{ik} x_{jk}\left(G_{k}-L_{k}\right) 
\label{equ_1}
\end{equation}
where $V_{I I}$ is the set of type II nodes, $r_{i j}$ is the mutual energy supply of node $i$ and node $j$, $s_{i j}$ is the sum of energy supply from the common neighbor nodes of node $i$ and node $j$.

\subsection{Constraints for type-III nodes}
For type-III nodes, according to the survivability standard of $N-3$, the current node is required to still work normally when the system loses 3 nodes arbitrarily. Similar to the situation of type-II nodes, when the worst scenario occurs, the power supply of the current node and the other two neighbor nodes are damaged. At this time, loads of the three nodes need to obtain sufficient support from the neighbor nodes. Therefore, a type-III node $j$ should satisfy the following constraints.

\begin{equation}
\begin{aligned}
S_{j}+S_{i}-r_{j i}-s_{j i}+S_{k}-r_{j k}-s_{j k}+t_{i j k} \\
-r_{i k}-s_{i k} \geq L_{i}+L_{j}+L_{k} \\ \forall j \in V_{I I I}, \forall i, k \in V, i \neq j \neq k \\
\end{aligned}
\end{equation}
\begin{equation}
t_{i j k}= \sum_{l=1, l \neq i, l \neq j, l \neq k}^{n} x_{i l} x_{j l} x_{k l}\left(G_{l}-L_{l}\right)
\label{equ_2}
\end{equation}
where $V_{I I I}$ is the set of type-III nodes, $t_{i j k}$ is the total power supply of common neighbor nodes of nodes $i$, $j$ and $k$.

\subsection{Objective function}

The objective of an MNSDP is to minimize the total length of the pre-set power supply circuits. It's worth mentioning that several different distances can be utilized, e.g., Manhattan distance, Chebyshev distance and Euclidean distance. However, different distances do not affect the subsequent calculation process and the algorithm framework, but only has a certain impact on the optimal result. Therefore, the Euclidean distance is adopted in this study. The objective function of the MNSDP can be expressed as follows:

\begin{equation}
Y=\text{min}\ \frac{\sum_{i,j=1}^{n}x_{ij}D_{ij}}{2}
\end{equation}
where $D_{ij}$ are distances between node $i$ and node $j$, $x_{ij}=1$ means there is a pre-set power supply circuit between node $i$ and node $j$. 

The decision variables of an MNSDP are represented as a symmetry binary matrix, which can be categorized as a BMOP. Since \eref{equ_1} and \eref{equ_2} contain product terms of boolean variables, this model belongs to mixed-integer nonlinear programming, which can be solved by commercial solver software \cite{bixby2012brief} like Gurobi \cite{bixby2007gurobi} and ILOG CPLEX \cite{manual1987ibm}. It's worth mentioning that, in this study, we only consider the situation when $K\in[1,2,3]$ for research purposes. For real-world problems, $K=3$ is enough to describe a high-reliability system. However, based on the model proposed in this work, the MNSDP model can be easily extended to other more complex situations.

Let's assume that for an MNSDP with $n$ nodes, where the numbers of type-I, type-II and type-III nodes are $n_1$, $n_2$ and $n_3$ respectively. Then, the number of constraints for these three types of nodes are $n_1$, $n_2*n$ and $n_3*n^2$. Specifically, for a 100-node system with 30, 30 and 40 type-I, type-II and type-III nodes, the constraints number is 403030. Intuitively, there are many constraints, which cause a huge challenge to seek for the global optimum.

\section{The proposed method}
\label{sec_algorithm}

Traditional EAs are effective in dealing with low-dimension problems. Most of them accept solutions with better constraint violations and objective values. Therefore, many potential solutions that are closer to the optimal will be discarded and the searching is trapped into local optima. In addition, to deal with the large-scale BMOPs, existing EAs are low-efficient by simulated binary crossover (SBX) and polynomial mutation (PM) operator. Therefore, an efficient EA that can handle large-scale decision variables is needed to address the BMOPs. 

Motivated by the differential evolution algorithms and the existing constrain-handling techniques, in this study, we proposed a novel improved feasibility rule based differential evolution algorithm for binary matrix optimization problems, termed LBMDE. To be specific, based on traditional differential evolution, a binary matrix-oriented operator is proposed that can accelerate the searching process. Moreover, to maintain the diversity of solutions and enhance the searching ability of the LBMDE, a novel environmental selection strategy based on an improved feasibility rule is adopted. The detail of the proposed LBMDE is illustrated in the following subsections. 

\subsection{Framework}

The framework of LBMDE is described in Algorithm \ref{alg_framework}. Similar to most EAs, LBMDE consists of the following parts: population initialization, offspring generation and environmental selection.

\renewcommand{\algorithmicrequire}{\textbf{Input:}} 
\renewcommand{\algorithmicensure}{\textbf{Output:}} 
\begin{algorithm}[htb]
\caption{General Framework of LBMDE}
\label{alg_framework}
\begin{algorithmic}[1]
\REQUIRE {Maximum generations $MaxGen$, population size $N$, problem parameters $ProPara$}
\ENSURE {Optimal solution $BestSol$}
\STATE $Pop \leftarrow Initialization(N,ProPara)$ /* Using heuristic method to initialize high-quality population */
\WHILE{$gen \leq MaxGen$}
\STATE $Off \leftarrow DEVariation(Pop)$ /* Special DE to generate new solutions */
\STATE $Pop \leftarrow EnvSel(Pop,Off,N)$ /* Improved feasible rule based method to select solutions */
\STATE $Pop \leftarrow Mutation(Pop)$ /* Utilize a mutation strategy to help converge */
\STATE $gen \leftarrow gen+1$
\ENDWHILE
\STATE $BestSol \leftarrow min(Pop)$
\end{algorithmic}
\end{algorithm}

As we can see from Algorithm \ref{alg_framework}, we first use a heuristic method to generate solutions with high quality ($line\ 1$), which can greatly accelerate the converging process. After that, a modified DE operator is performed to produce offspring, which is designed for BMOPs, see $line\ 3$. In addition, an improved feasible rule based environmental selection method is adopted to enhance the diversity of solutions and handle the constraints. Moreover, once the evolution of the population is at a standstill, a mutation operator will be conducted to help explore the decision space, see $line\ 5$. The detailed information about each step will be described in the following subsections.

\subsection{General initialization method for BMOPs}
\label{sec_init}
For the BMOP studied in this work, the number of decision variables is huge. For a system with $n$ nodes, the number of valid decision variables is $\frac{n^2-n}{2}$. In addition, since the objective is to minimize the total length of the power supply circuits, many decision variables should be set to 0. Therefore, the MNSDP is a typical large-scale sparse optimization problem \cite{kropp2022benefits}. As a result, to accelerate the searching process, it's quite necessary to generate high-quality solutions through heuristic methods. However, it's a tough task to balance the diversity and the quality of solutions. Once the heuristic method can not provide diverse solutions, the searching will easily get trapped into local optima \cite{burke1998initialization}. For large-scale sparse problems, this could be fatal since it's hard to get rid of the local optima through crossover and mutation \cite{tian2019evolutionary}.

\begin{algorithm}[phb]
\caption{Heuristic method}
\label{alg_init}
\begin{algorithmic}[1]
\REQUIRE {Number of nodes $n$, distance between all nodes $D$}
\ENSURE {High-quality solution $X$}
\STATE $X \leftarrow GenerateMatrix(n)$ /* Set each decision variable in $X$ equals to 1 */
\STATE $count \leftarrow 0$
\WHILE {$count<n$}
\STATE $X1 \leftarrow X$
\STATE $p \leftarrow RandomPick(n)$ /* Randomly select a node */
\STATE $remainP \leftarrow FindNeighbor(p)$
\STATE $selP \leftarrow RouletteWheelSel(D(p,remainP))$ /* Randomly select a node according to \eref{equ_3} */
\STATE $X1(p,selP) \leftarrow 0$
\IF {$isfeasible(X1)$}
\STATE $X \leftarrow X1$
\STATE $count \leftarrow 0$
\ELSE
\STATE $count \leftarrow count+1$ /* The counter is used to evaluate if $X$ can be improved or not */
\ENDIF
\ENDWHILE
\end{algorithmic}
\end{algorithm}

To this end, we proposed a general heuristic framework to generate high-quality solutions for BMOPs, which can be seen in Algorithm \ref{alg_init}. In the beginning, the solution is initialized with all decision variables equal to 1, see $line\ 1$. Then during each iteration, we randomly select a connection between two different nodes and delete it, see $lines\ 4-8$. If the new solution is feasible, then we accept it as a new solution with better quality. Otherwise, the counter will be added by one. Then, once the counter is larger than a threshold value, e.g., the total number of nodes $n$ in this work, we consider there is no room for improvement and the current network structure is considered a high-quality initialized solution. Specifically, the second node is selected through the roulette wheel selection \cite{lipowski2012roulette} according to the probabilities. The probability $p_i$ for selecting the neighbor node $i$ of node $j$ can be expressed as follows:

\begin{equation}
p_i=\frac{D_{ij}}{\sum_{k\in N_{j}}^{} D_{jk}}
\label{equ_3}
\end{equation}
where $D_{ij}$ and $N_{j}$ are the distance between node $i$, $j$ and the neighbor nodes set of node $j$.

It's worth mentioning that, the heuristic method used here is designed for minimization problems. However, since this heuristic method is parameter-free, it can be extended to initialize solutions for other BMOPs, e.g., the CNN structure design problems \cite{li2017structural}.

\subsection{Binary-matrix-based DE operator for BMOPs}
There are many variants of DE operators that can be described using the notation $DE/x/y/z$ \cite{opara2019differential,lin2011comparative}, where $x$ is the vector to be mutated, $y$ is the number of difference vectors, and $z$ denotes the crossover scheme. Specifically, for the classic DE algorithm of $DE/rand/2/bin$, we assume a population consists of $N$ solutions $\mathbf{x}_{i,G}$, $i=1,2,...,N$ for each generation $G$. Then the mutation can be described as follows:
\begin{equation}
\mathbf{v}_{i,G+1}=\mathbf{x}_{r1,G}+F \cdot (\mathbf{x}_{r2,G}-\mathbf{x}_{r3,G})
\label{equ_de}
\end{equation}
where $r1$, $r2$ and $r3$ are randomly selected solution index, $F \in [0,2]$ is a real and constant factor that controls the amplification of the difference vector $(\mathbf{x}_{r2,G}-\mathbf{x}_{r3,G})$.

For the crossover operator of the DE algorithm, we assume $\mathbf{u}_{ji,G}$ to be the $j$-th decision variable in generation $G$ of the solution $\mathbf{x}_i$. Then, the updated process of DE can be expressed as:
\begin{equation}
\mathbf{u}_{j i, G+1}= \begin{cases}\mathbf{v}_{j i, G+1} & \text { if }(r(j) \leq C R) \text { or } j=r n(i), \\ \mathbf{u}_{j i, G} & \text { if }(r(j)>C R) \text { and } j \neq r n(i)\end{cases}
\end{equation}
where $CR\in [0,1]$ means the crossover constant. $r n(i)\in(1,2,...,D)$, $D$ is number of decision variables, is a randomly chosen index which ensures that $\mathbf{u}_{j i, G+1}$ gets at least one element from $\mathbf{v}_{i, G+1}$. It's worth mentioning, $\mathbf{v}_{i, G+1}$ could be randomly selected from the current population or the best solution.

\begin{figure}[tbph]
	\begin{center}
		\includegraphics[width=3.5in]{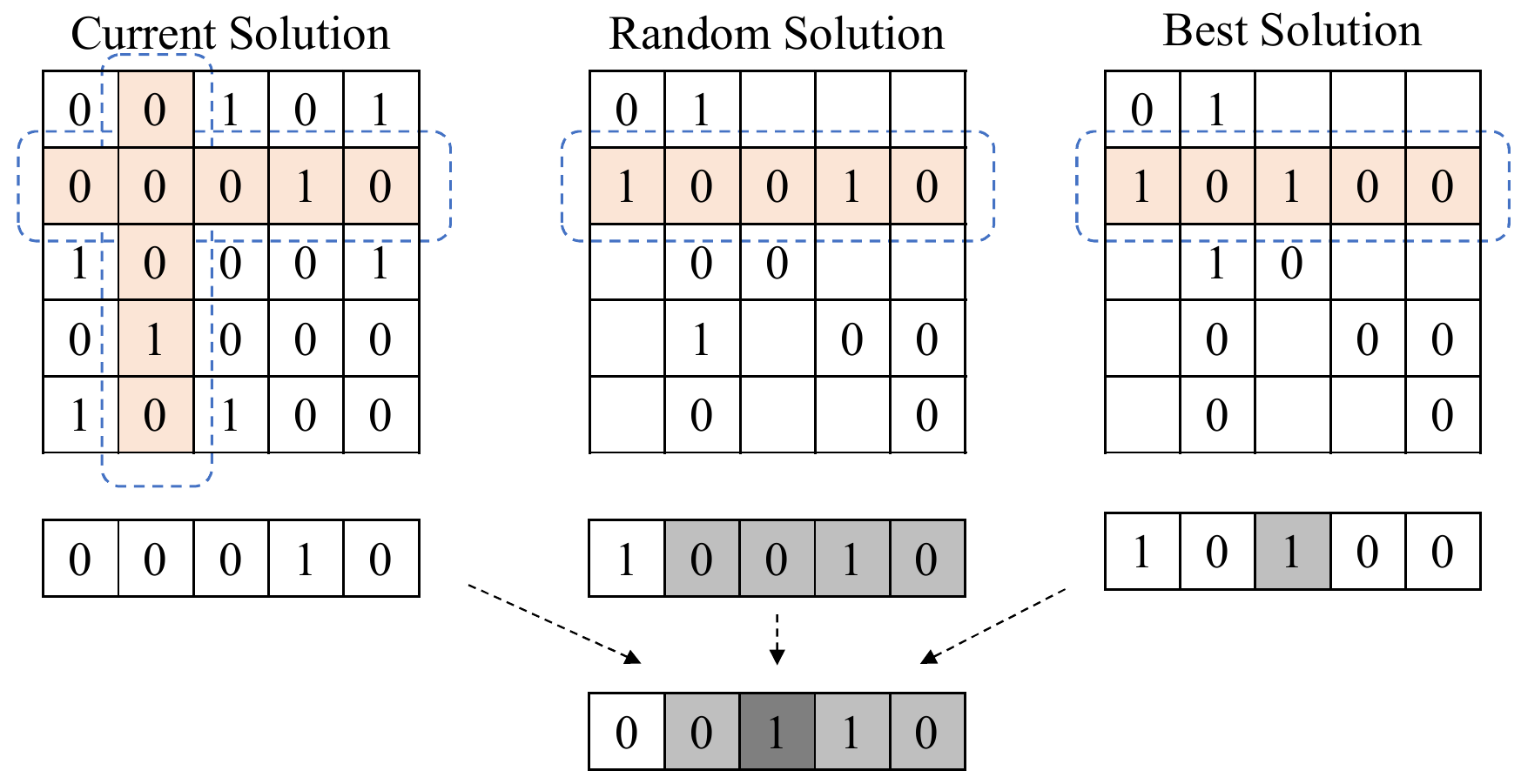}
		\caption{Schematic diagram of the proposed DE-based binary matrix crossover operator.}
		\label{fig_operator}
	\end{center}
\end{figure}

Traditional crossover and mutation operators randomly select a certain number of gene points. Such a method is effective in solving problems with low-dimension decision variables. However, for large-scale problems, such a method could be low-efficient. To address this problem, we proposed a binary-matrix-based DE operator for BMOPs. Specifically, the $k$-th decision variable in the $j$-th row/column in generation $G$ is updated according to the following equation:

\begin{equation}
\mathbf{x}_{G+1}^{j,k}= \begin{cases}
\mathbf{best}_{G+1}^{j,k} & \text { if }(r1(k) \leq F), \\ \mathbf{u}_{G+1}^{j,k} & \text { if }(r1(k)>F\ \text{and}\ r2(k)\leq CR),
\\ \mathbf{x}_{i, G}^{j,k} & \text { else }
\end{cases}
\end{equation}
where $\mathbf{best}$ and $\mathbf{u}$ are the global best solution and a randomly selected solution. $r1(k),r2(k)\in (0,1)$ are random numbers.

The solution updating process is presented in \fref{fig_operator}. To be specific, we first select the second row of the decision variables, which contains 5 binary values in this case. Then, four (randomly selected according to $F$) gene points in the random solution $\mathbf{u}$ and one (randomly selected according to by $CR$) gene point in the best solution $\mathbf{best}$ are picked to update the current solution. Therefore, the gene of the $j$-th row/column consists of three parts: the current solution, the randomly selected solution, and the best solution. 

Notably, in a crossover procedure, several rows/columns will be selected with probability $1/n$, which is set according to the suggested parameter used in SBX and PM. By adopting the novel DE operator, the current solution can obtain high-quality gene fragments from other solutions. Moreover, since this information is provided with rows/columns (in this study, it means the neighborhood structure), the connection information can be retained to a great extent. Therefore, the search process can be accelerated enormously.

\subsection{Environmental selection}
\label{sec_sel}
Generally, balancing the convergence quality in the objective and the constraints is challenging. Therefore, many constraint-handling techniques have been proposed to enhance the searching ability of EAs, e.g, feasible rules \cite{mezura2004simple}, $\epsilon$-constrained method \cite{takahama2010efficient}, penalty function \cite{yeniay2005penalty}, stochastic ranking \cite{runarsson2000stochastic}, multi-objective optimization-based method \cite{ray2009infeasibility} and other hybrid techniques. Among them, the feasible rules are effective and widely adopted. To be specific, solution $\mathbf{x}_i$ is said to be better than solution $\mathbf{x}_j$ if any of the following conditions is true:
\begin{itemize}
\item Solution $\mathbf{x}_i$ is feasible while solution $\mathbf{x}_j$ is not.
\item Solutions $\mathbf{x}_i$ and $\mathbf{x}_j$ are feasible, and the objective value of $\mathbf{x}_i$ is better than $\mathbf{x}_i$.
\item Solutions $\mathbf{x}_i$ and $\mathbf{x}_j$ are infeasible, and the constraint violation of  $\mathbf{x}_i$ is smaller than $\mathbf{x}_j$.
\end{itemize}

By adopting the feasible rules, EAs will select solutions with smaller constraint violations and better objective values. As we can observe, solutions with better objective value but worse constraint violation can not be retained. However, for these solutions, they may be closer to the true optimal since their objective values are smaller. As a result, algorithms will easily get trapped into local optima. To address this issue, many studies have been carried out. Wang et al. \cite{wang2015incorporating} proposed to incorporate objective function information into the feasibility rule. Tian et al. \cite{tian2020coevolutionary} introduced a coevolutionary framework that two populations focus on different goals. One is to find the true Pareto front without considering constraints, another is to find feasible solutions by feasible rules. Then, in \cite{liu2019handling}, ToP is proposed where the first stage is to find promising feasible regions and the second stage is to obtain the final solutions. Recently, Ming et al. \cite{ming2021dual} introduced a cooperative coevolutionary algorithm that maintains two collaborative populations by utilizing the self-adaptive penalty function and the feasibility-oriented approach. In short, accepting solutions with worse constraint violation but better objective value can significantly enhance the performance of EAs.

\renewcommand{\algorithmicrequire}{\textbf{Input:}} 
\renewcommand{\algorithmicensure}{\textbf{Output:}} 
\begin{algorithm}[!thb]
\caption{Environmental Selection}
\label{alg_selection}
\begin{algorithmic}[1]
\REQUIRE {Population $Pop$, offspring $Off$, population size $N$}
\ENSURE {Updated Population $Pop$}
\STATE $[PopCon,OffCon] \leftarrow GetCon(Pop,Off)$ /* Get the constraint violations of all solutions */
\STATE $[PopObj,OffObj] \leftarrow GetObj(Pop,Off)$ /* Get the objective values of all solutions */
\STATE $next \leftarrow PopCon>OffCon$
\STATE $Pop(next) \leftarrow Off(next)$ /* If the constraint violation of offspring solution is less than that of its parent, then replace its parent */
\STATE $remain \leftarrow PopCon<OffCon \& PopObj>OffObj$
\STATE $Arc \leftarrow Off(remain)$ /* Solutions with better objective values but worse constraint violations may be potential good solutions */
\FOR{$i=1:N/batchsize$}
\STATE $pick \leftarrow rand(N,batchsize)$ /* Randomly select $batchsize$ solutions */
\STATE $worstsol \leftarrow max(GetCon(Pop(pick)))$
\STATE $bestsol \leftarrow min(GetCon(Arc(pick)))$
\IF {$GetObj(worstsol)>GetObj(bestsol)$} 
\STATE $Pop(worst)\leftarrow Arc(best)$ /* Replace the worst solution with the best solution */
\ENDIF
\ENDFOR
\end{algorithmic}
\end{algorithm}

Then, borrowing the idea from the above-mentioned algorithms \cite{wang2015incorporating,tian2020coevolutionary,ming2021dual}, we propose an improved feasible rule based environmental selection strategy, which can be illustrated in Algorithm \ref{alg_selection}. The main idea of the proposed strategy is to increase the probability of an infeasible solution with a better objective value being selected. Therefore, there are roughly two stages. The first stage is to select solutions with smaller constraint violations and better objective values, which is performed according to the basic feasible rules, see $line\ 3-4$. The second stage aims to select some potential optimal solutions. First, solutions with better objective values but worse constraint violations will be stored in an archive ($line\ 5-6$). Then, for each batch of solutions, we randomly select some solutions from the archive and the main population. If the objective value of the best solution in the archive is better than the worst solution in the main population, the latter solution will be replaced by the former, see $line\ 9-13$. In this way, solutions that may be closer to the true optima will not be directly removed during the evolution. Therefore, the EAs can better get rid of local optima. 

\section{Experiment}
\label{sec_exp}
\subsection{Experimental setting}

\subsubsection{Benchmark Problems}
\label{sec_benchmark}
Since the mathematical model of the MNSDPs is firstly proposed in this study, there are no benchmark test suites that can be utilized to examine the performance of EAs. Therefore, in this study, an MNSDP library (MNSDP-LIB) is proposed which is adjustable in both scale and the solving difficulties. Specifically, there are two control parameters, e.g., the number of nodes $n$ and the ratio of energy generation and consumption $r$. The steps for generating MNSDP-LIB test instances can be described as follows: 1) in a cartesian coordinate system ($x,y\in[0,10]$), randomly pick $n$ points, which are used as the position of nodes; 2) randomly assigned a real value to each node as its energy generation; 3) the energy consumption of nodes can be calculated by its energy generation and parameter $r$; 4) $K_i$ for node $i$ is randomly picked from $[\{1,2,3\}$. As we can see, the number of constraints is highly related to the number of nodes with $N-3$. In addition, the smaller value of $r$, the more complex the network structure is (there should be more neighbors for nodes with smaller $r$). Therefore, the solving difficulty can be controlled by these two parameters.

Following the above construction method, we generated 25 test instances to comprehensively examine the performance of algorithms. To be specific, $n\in \{10,20,50,80,100\}$ and $r\in \{1.3,1,4,1,5,1,6,1.7\}$ are adopted. For the convenience of researchers, MNSDP-LIB is open-access \footnote{The test instances of MNSDP-LIB can be reached from \url{https://github.com/Wenhua-Li/LBMDEforMNSDP/tree/main/MNSDP-LIB}.}.

\subsubsection{Competitor Algorithms}
To verify the effectiveness of LBMDE in solving large-scale BMOPs, self-adaptive binary differential evolution algorithm (SabDE) \cite{BANITALEBI2016487}, matrix-binary codes based genetic algorithm (MGA) \cite{patle2018matrix}, Jaya-based binary optimization algorithm (JayaX) \cite{aslan2019jayax}, and binary particle swarm optimization (BPSO) \cite{ji2020bio} are chosen as competitor algorithms. Specifically, these algorithms are representative EAs designed for large-scale binary optimization problems based on different state-of-the-art EAs. Among them, SabDE is designed for large-scale binary optimization problems, MGA is proposed to solve BMOPs, JayaX and BPSO are utilized to better solve binary optimization problems with different solution updating methods. In addition, since the mathematical model of the MNSDP in this study can be easily transformed into a MIP problem that can be solved by commercial solvers, IBM ILOG CPLEX is adopted as the baseline.

For all the algorithms, we set the population size $N=20*n$, and the maximum number of function evaluations $N_E$ is set to $n*N$, where $n$ is the number of nodes. It's worth mentioning that, for algorithms that are unable to solve constraint optimization problems, the basic feasible rule based method is adopted. In addition, for a fair comparison, the population for all compared algorithms is initialized by the method proposed in \sref{sec_init}. All experiments are implemented on a PC configured with an Intel i9-9900X @ 3.50 GHz and 64 G RAM. For the convenience of subsequent researchers, the source code of LBMDE is open to access \footnote{The source code of LBMDE can be accessed from \url{https://github.com/Wenhua-Li/LBMDEforMNSDP}.}.

\subsection{Result analysis}
\subsubsection{Performance comparison}
\begin{figure}[tbph]
	\begin{center}
		\includegraphics[width=3.5in]{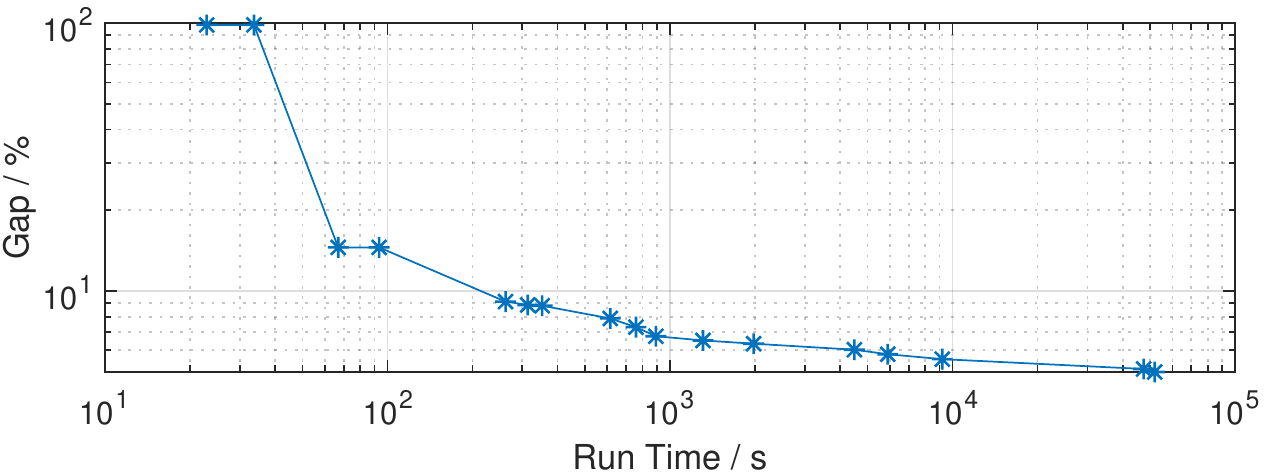}
		\caption{Convergence process over time of CPLEX on solving MNSDP-LIB with 100 nodes, where logarithmic axis is adopted in both x- and y-axis.}
		\label{fig_cplex}
	\end{center}
\end{figure}

This section shows the performance of LBMDE and the competitor algorithms on MNSDP-LIB. Specifically, 25 instances with a different number of nodes are adopted. The results are presented in \tref{tab_result}, where the mean and standard variance of the obtained objective values over 30 independent runs are listed. In addition, \fref{fig_runtime} shows the run time of all compared algorithms on problems with different numbers of nodes. It's worth mentioning that, for CPLEX, the running time for obtaining the final optimal results for problems with 80 nodes is unbearable (two to four days). Therefore, we set the gap to 3\%, which means that the current obtained solution is worse than the optimal solution within 3\% in terms of objective value. In addition, for problems with 100 nodes, gap = 5\% is adopted. \fref{fig_cplex} presents the searching process of CPLEX over time, from which we can see that the gap can be greatly decreased in the early phase (when gap $\geq 10\%$). However, during the later period, a small improvement in objective value requires a huge time cost.

\begin{table*}

\caption{Average and standard variance results of the compared algorithms on MNSDP-LIB test suite, where the best mean for each test instance is highlighted.}
\centering
\begin{tabular}{ccccccc}
\hline
\textbf{Problems} & \textbf{CPLEX} & \textbf{SabDE} & \textbf{MGA} & \textbf{JayaX} & \textbf{BPSO} & \textbf{LBMDE} \bigstrut\\
\hline
\textbf{MNSDP-10-1} & 181.40  & 184.20(1.78) & 182.75(1.19) & 194.96(9.31) & 194.80(7.41) & \textbf{181.53(1.00)} \bigstrut[t]\\
\textbf{MNSDP-10-2} & 141.99  & 148.69(3.61) & \textbf{144.11(2.04)} & 157.76(6.66) & 162.18(9.23) & 144.29(2.38) \\
\textbf{MNSDP-10-3} & 122.52  & 126.90(2.84) & 123.01(0.99) & 139.75(5.13) & 137.52(8.11) & \textbf{122.65(0.44)} \\
\textbf{MNSDP-10-4} & 123.45  & 129.76(4.85) & \textbf{124.36(1.96)} & 151.60(4.72) & 143.14(7.34) & 124.65(2.48) \\
\textbf{MNSDP-10-5} & 113.38  & 116.44(4.17) & 114.42(1.75) & 132.13(4.83) & 131.64(6.99) & \textbf{113.48(0.20)} \\
\textbf{MNSDP-20-1} & 258.61  & 289.68(10.13) & 278.29(4.43) & 376.35(24.51) & 334.99(18.08) & \textbf{260.44(2.30)} \\
\textbf{MNSDP-20-2} & 184.49  & 205.67(7.59) & 192.91(1.79) & 268.49(21.65) & 243.05(14.80) & \textbf{190.47(2.77)} \\
\textbf{MNSDP-20-3} & 161.47  & 179.47(6.49) & 161.50(0.17) & 269.58(21.48) & 227.34(17.48) & \textbf{161.95(0.98)} \\
\textbf{MNSDP-20-4} & 142.58  & 157.33(6.77) & 144.10(1.30) & 211.16(12.09) & 185.12(10.87) & \textbf{144.36(1.44)} \\
\textbf{MNSDP-20-5} & 129.96  & 143.19(6.63) & 130.74(0.75) & 200.66(13.69) & 178.02(13.83) & \textbf{130.92(0.53)} \\
\textbf{MNSDP-50-1} & 375.95  & 483.19(18.36) & 429.01(4.19) & 608.31(35.67) & 635.11(137.90) & \textbf{387.39(2.39)} \\
\textbf{MNSDP-50-2} & 263.97  & 351.72(17.90) & 303.57(3.38) & 444.83(20.24) & 432.63(20.02) & \textbf{266.29(0.72)} \\
\textbf{MNSDP-50-3} & 246.66  & 338.29(19.58) & 281.65(3.70) & 419.88(18.83) & 494.43(184.09) & \textbf{250.41(1.26)} \\
\textbf{MNSDP-50-4} & 199.91  & 290.04(17.41) & 239.98(4.16) & 352.80(18.98) & 708.93(70.51) & \textbf{203.71(0.66)} \\
\textbf{MNSDP-50-5} & 188.39  & 284.43(16.77) & 232.42(4.54) & 339.80(17.15) & 676.48(59.46) & \textbf{190.96(0.57)} \\
\textbf{MNSDP-80-1} & 418.99(3\%) & 686.82(39.19) & 536.83(5.84) & 787.28(37.47) & 2728.84(203.14) & \textbf{436.33(2.80)} \\
\textbf{MNSDP-80-2} & 330.32(3\%) & 572.95(37.23) & 474.60(6.17) & 676.01(27.23) & 684.80(20.75) & \textbf{348.42(3.15)} \\
\textbf{MNSDP-80-3} & 286.70(3\%) & 500.67(33.57) & 397.96(5.60) & 579.95(31.38) & 2625.37(424.05) & \textbf{301.66(2.70)} \\
\textbf{MNSDP-80-4} & 252.89(3\%) & 477.30(25.79) & 371.58(6.00) & 555.20(18.73) & 2659.89(167.35) & \textbf{267.46(1.70)} \\
\textbf{MNSDP-80-5} & 225.22(3\%) & 432.58(18.52) & 330.15(5.71) & 514.41(27.30) & 2615.20(168.24) & \textbf{239.71(2.46)} \\
\textbf{MNSDP-100-1} & 456.70(5\%) & 817.28(35.32) & 718.46(6.69) & 926.81(43.67) & 971.26(40.39) & \textbf{506.77(9.33)} \\
\textbf{MNSDP-100-2} & 368.34(5\%) & 675.98(32.54) & 564.73(7.48) & 791.71(38.84) & 4363.96(1632.37) & \textbf{397.42(6.76)} \\
\textbf{MNSDP-100-3} & 298.45(5\%) & 608.29(39.66) & 462.56(5.04) & 705.41(26.56) & 4847.05(274.28) & \textbf{321.89(4.16)} \\
\textbf{MNSDP-100-4} & 275.67(5\%) & 602.28(45.35) & 426.05(6.01) & 692.19(25.16) & 5082.35(278.91) & \textbf{295.19(2.54)} \\
\textbf{MNSDP-100-5} & 239.65(5\%) & 556.89(21.67) & 400.08(5.62) & 634.38(25.63) & 4965.50(310.12) & \textbf{264.09(4.56)} \bigstrut[b]\\
\hline
\end{tabular}%

\label{tab_result}
\end{table*}

\tref{tab_result} shows the average objective values comparison results, from which we can observe that LBMDE shows better performance than other state-of-the-art EAs on the chosen test problems. Specifically, among all EAs, LBMDE wins 23 instances over 25 problems. In addition, MGA obtains the best result for MNSDP-LIB-10-2 and MNSDP-LIB-10-4. Compared to the baseline (true optimal solution obtained by the MIP method), results obtained by LBMDE are still competitive. For small-scale problems ($n\leq20$), LBMDE can obtain stable results with good quality. To be specific, the gap between the final results of LBMDE and the true optimal is less than 1\% for most of the test instances while the variance is small. For large-scale problems, LBMDE still shows its advantage. For some problems with 80 nodes, results obtained by LBMDE are better than the baseline (the baseline is obtained with setting gap=3\% for problems with 80 nodes). In addition, LBMDE shows high stability in dealing with large-scale problems.

As we can see from \tref{tab_result}, except for LBMDE, MGA and SabDE show satisfactory performance in solving small-scale problems. For problems with 10 or 20 nodes, the number of decision variables is 45 and 190 respectively, which is relatively small. Therefore, the existing EAs can somehow figure out the optimal solutions for these problems. However, for problems with large-scale decision variables, these algorithms show poor searching efficiency compared to LBMDE. The BPSO shows the worst performance over the compared algorithms. In BPSO, the real values are utilized to encode a solution. Further study shows that the velocities of particles in BPSO tend to be zero. Problems studied in this work can be categorized as large-scale sparse optimization problems. Therefore, most of the decision variables should be zero. For BPSO, the updating process of particle velocity is related to the particle position, which is getting smaller as the evolution goes on. Therefore, BPSO will quickly converge to a local optimum and standstill.

\begin{figure*}[htbp]
\setcounter{subfigure}{0}
\centering
\subfigure[CPLEX]{\includegraphics[width=1.7in]{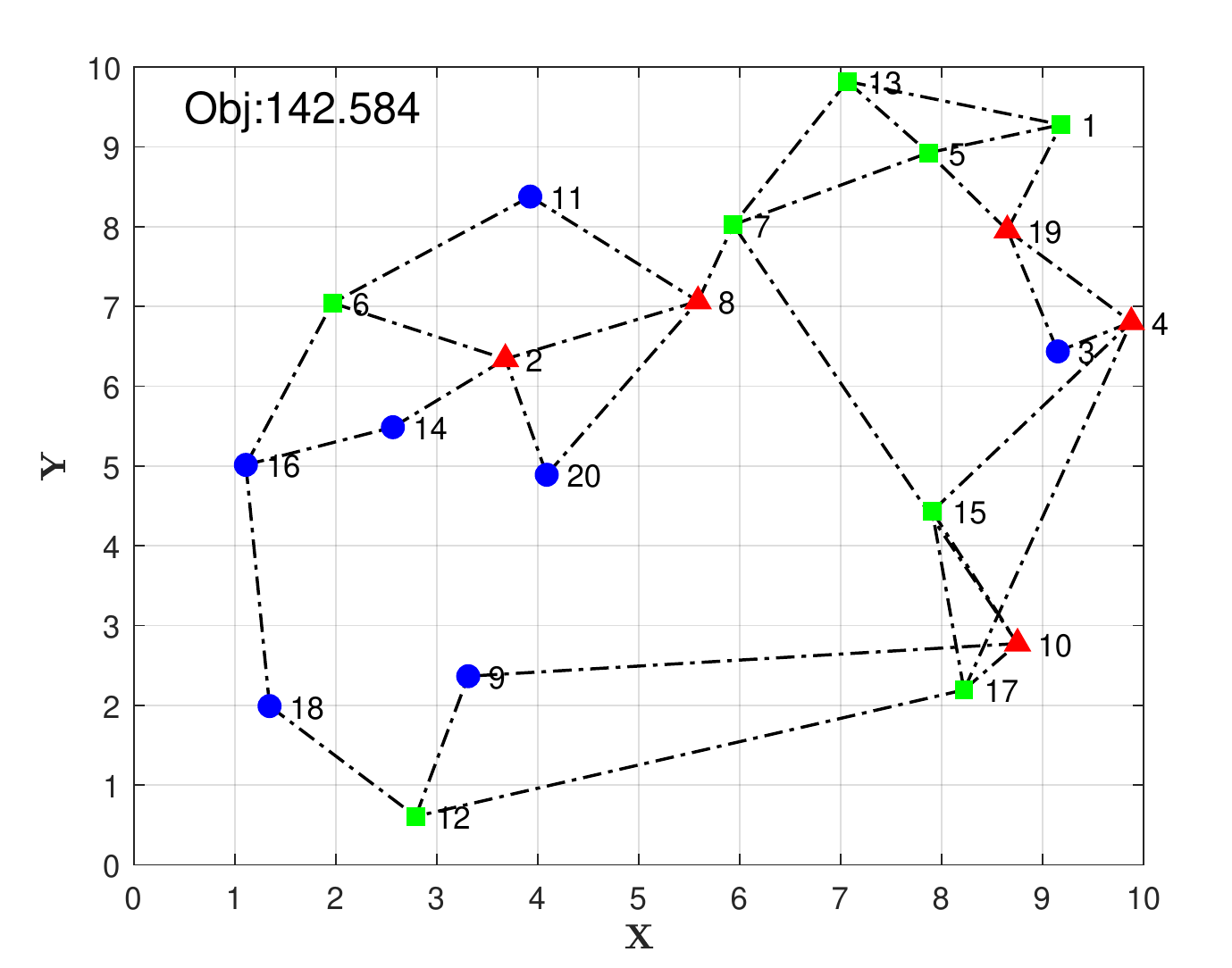}}
\subfigure[SabDE]{\includegraphics[width=1.7in]{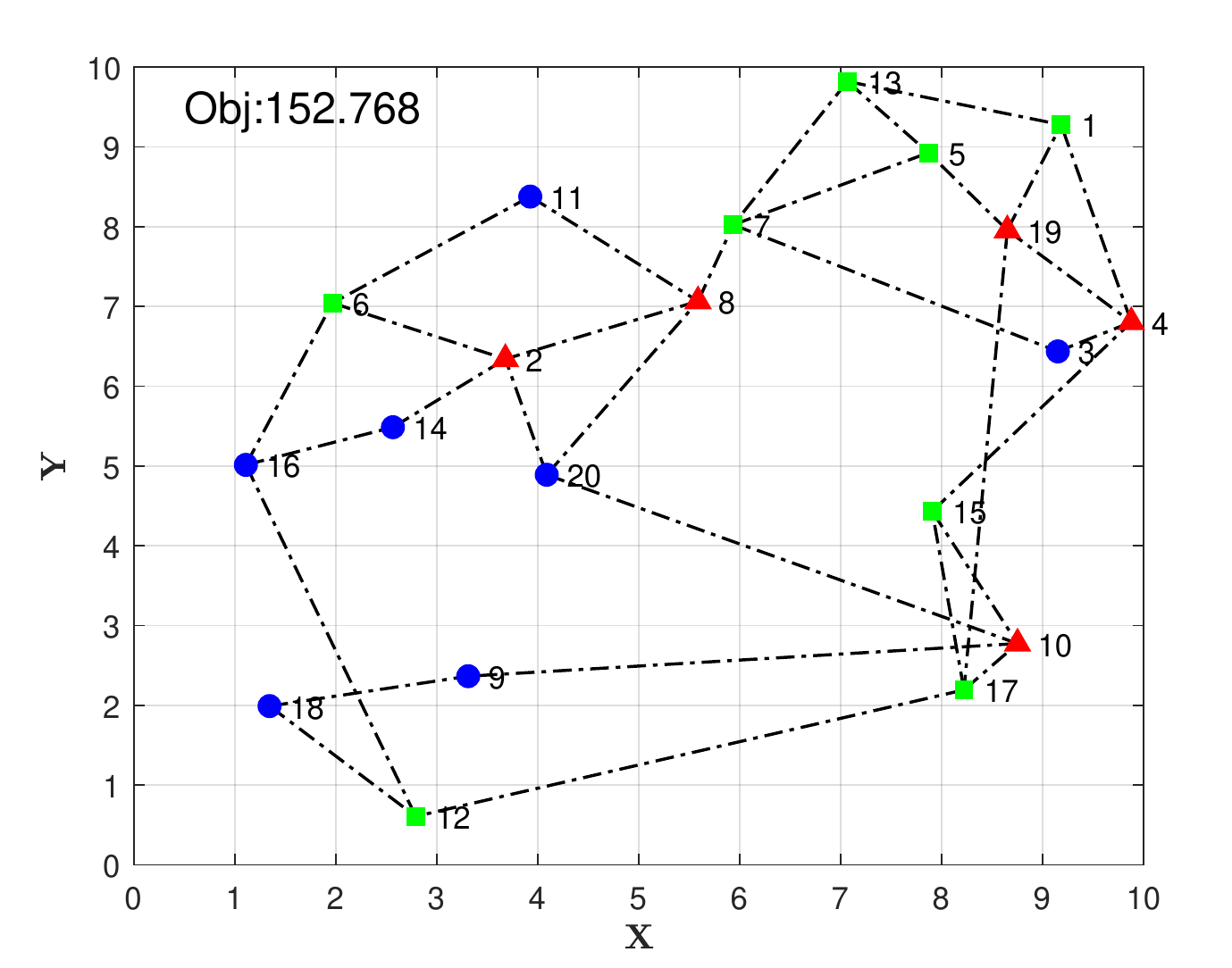}}
\subfigure[MGA]{\includegraphics[width=1.7in]{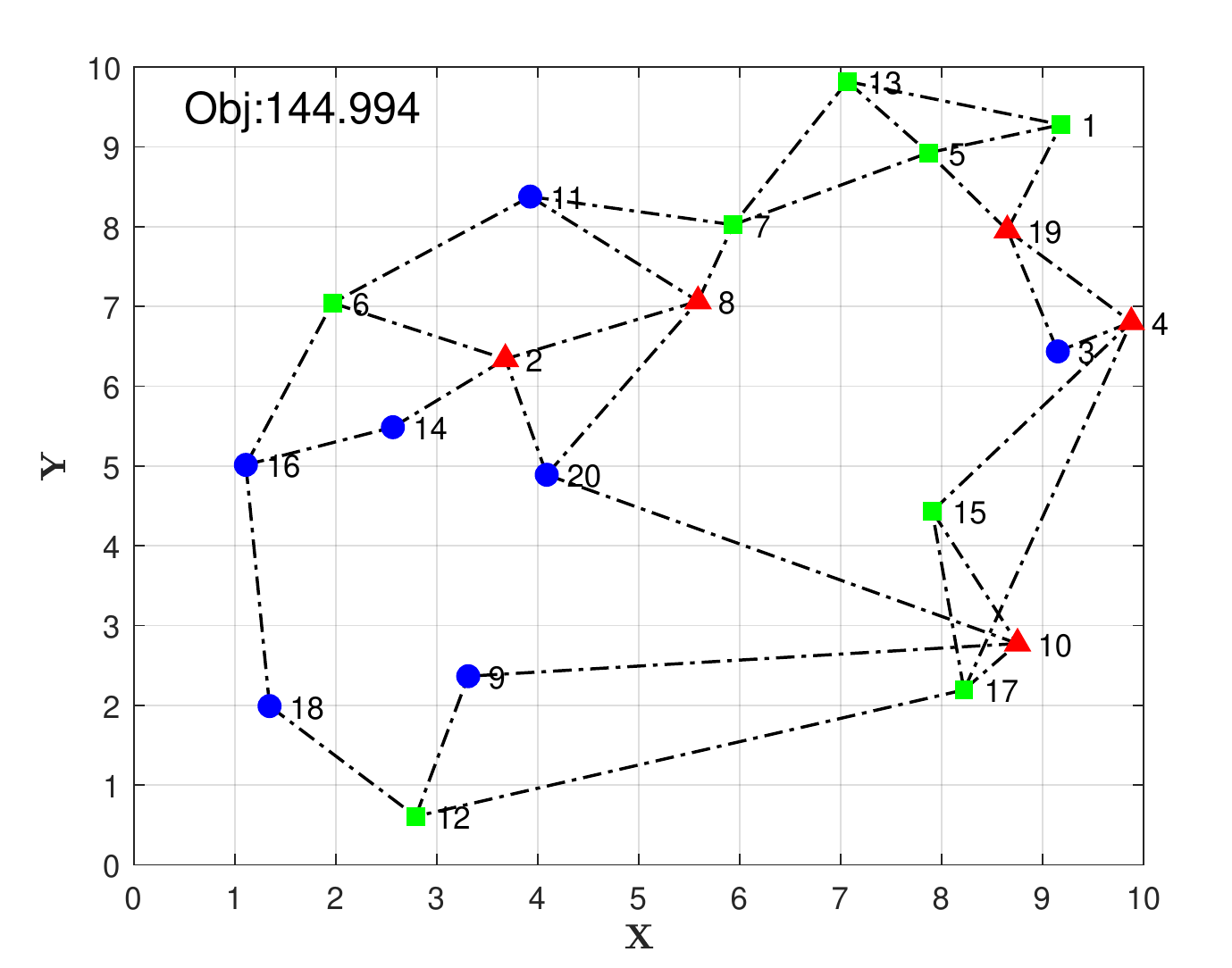}}
\subfigure[JayaX]{\includegraphics[width=1.7in]{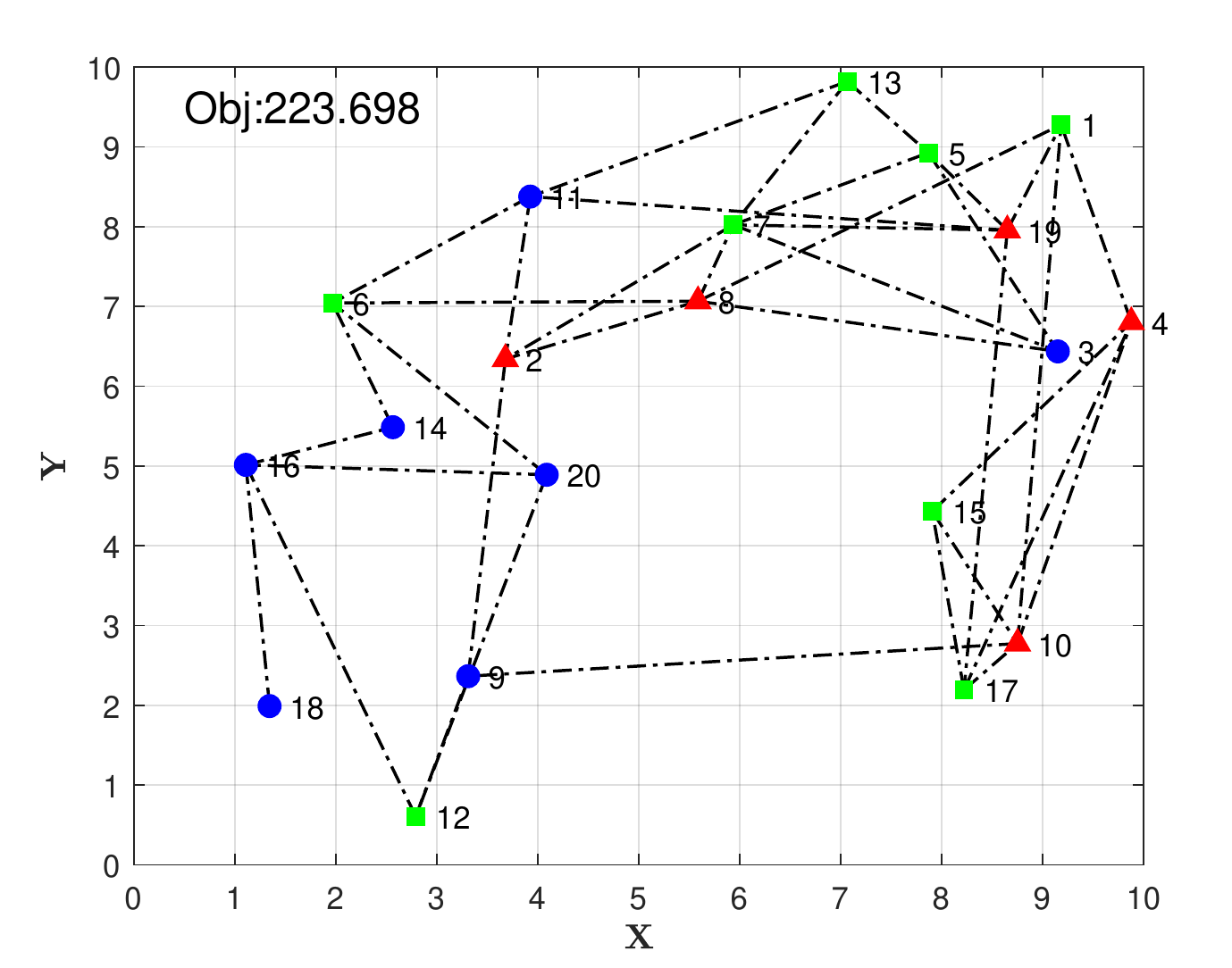}}
\subfigure[BPSO]{\includegraphics[width=1.7in]{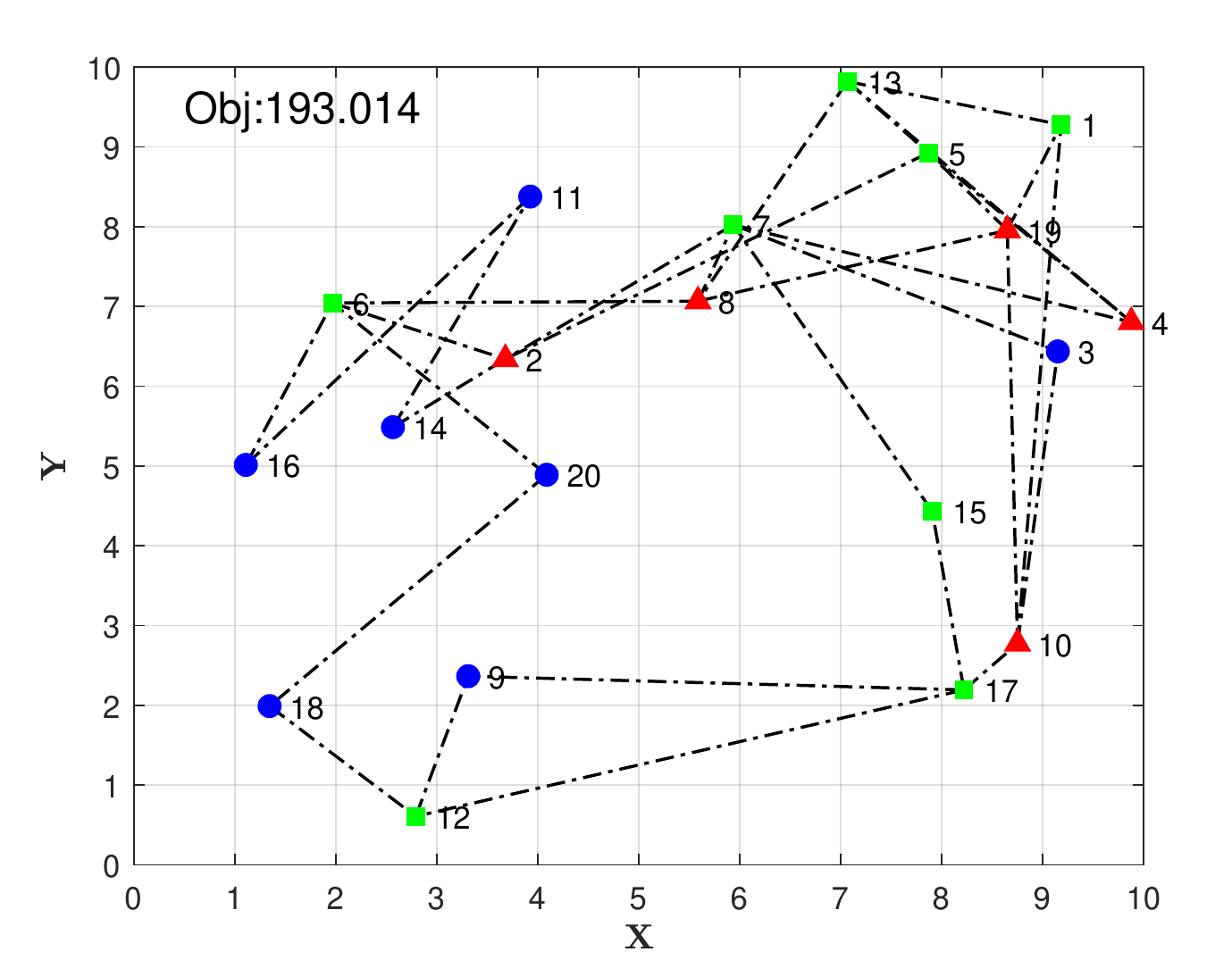}}
\subfigure[LBMDE]{\includegraphics[width=1.7in]{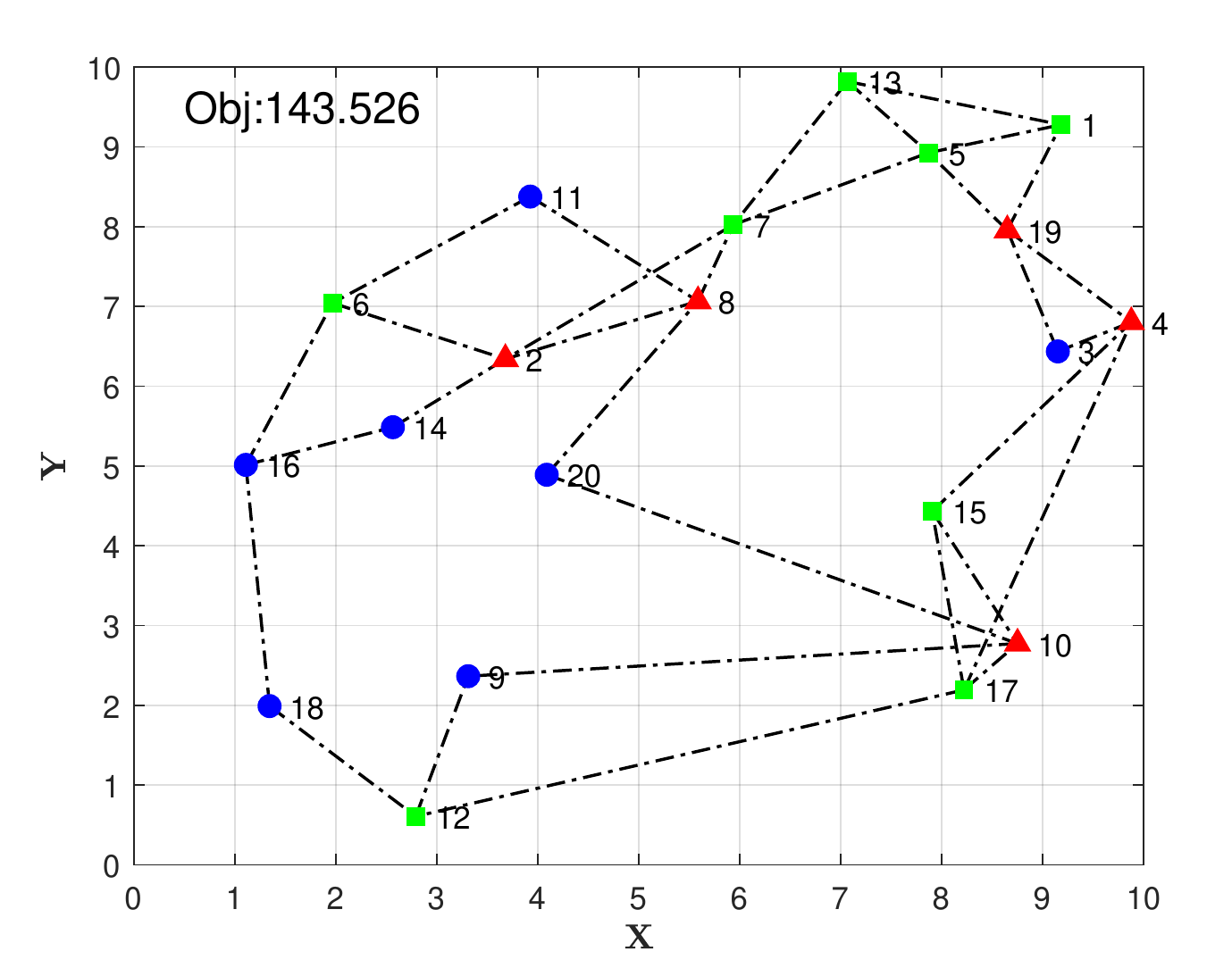}}
\subfigure[CPLEX]{\includegraphics[width=1.7in]{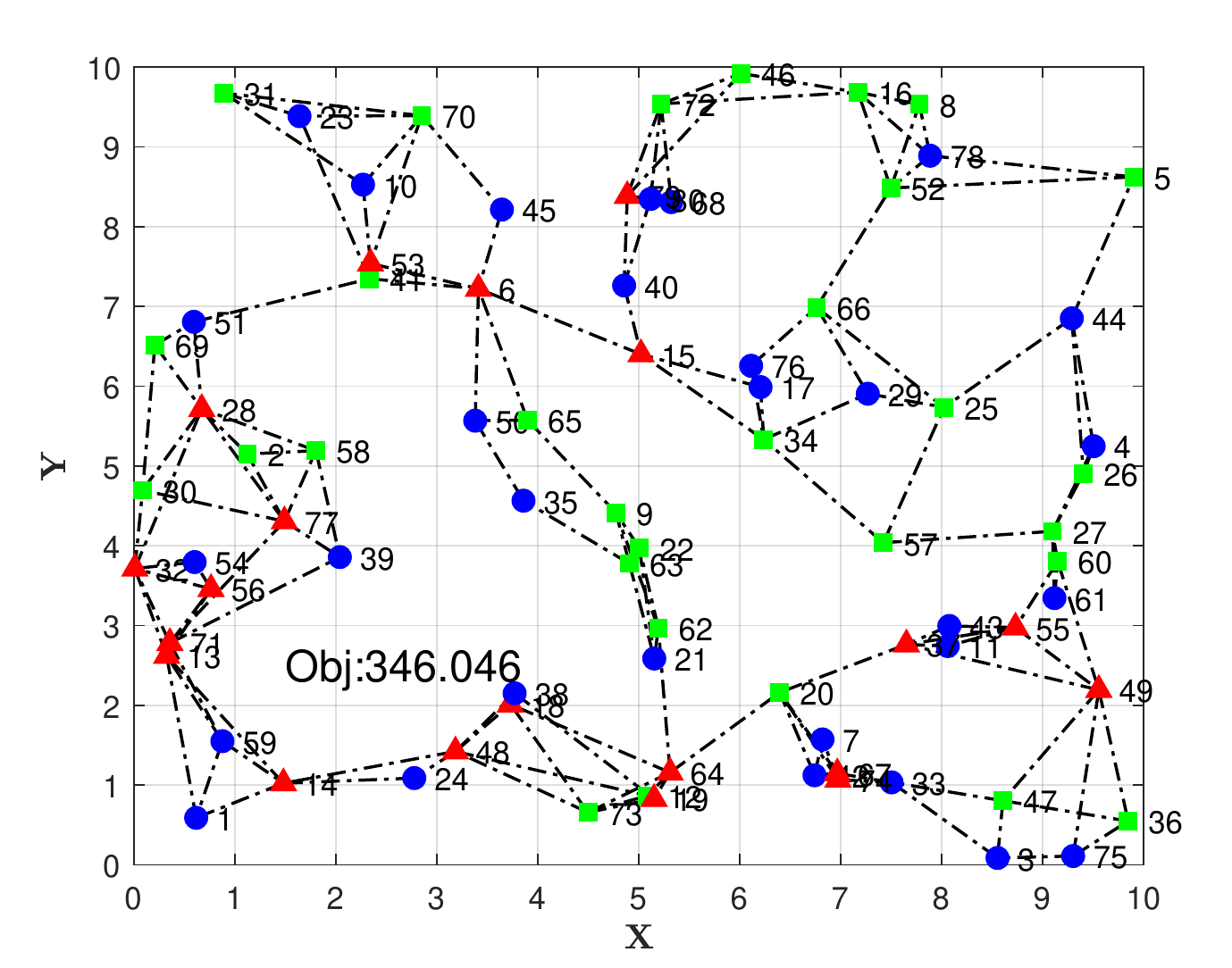}}
\subfigure[SabDE]{\includegraphics[width=1.7in]{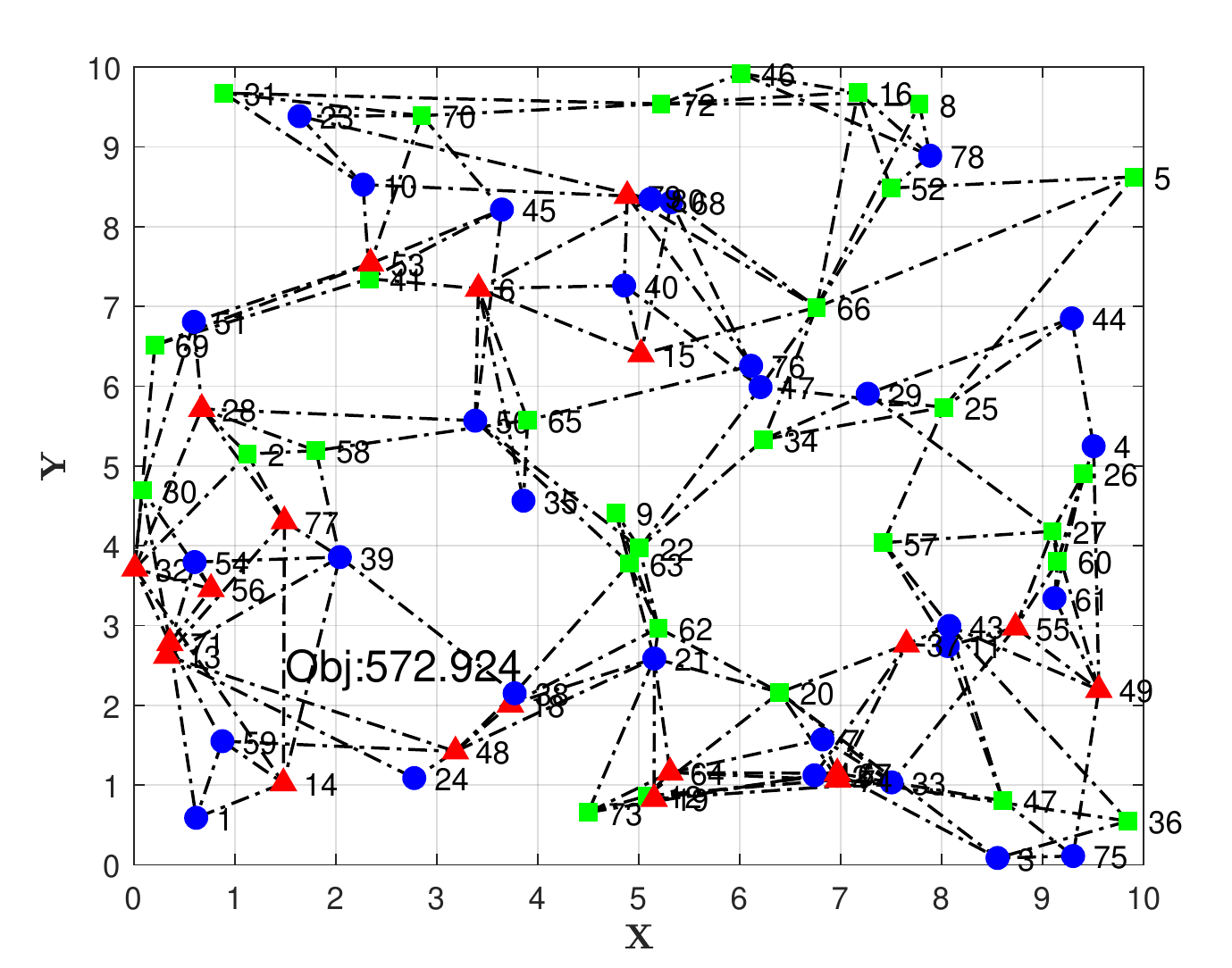}}

\subfigure[MGA]{\includegraphics[width=1.7in]{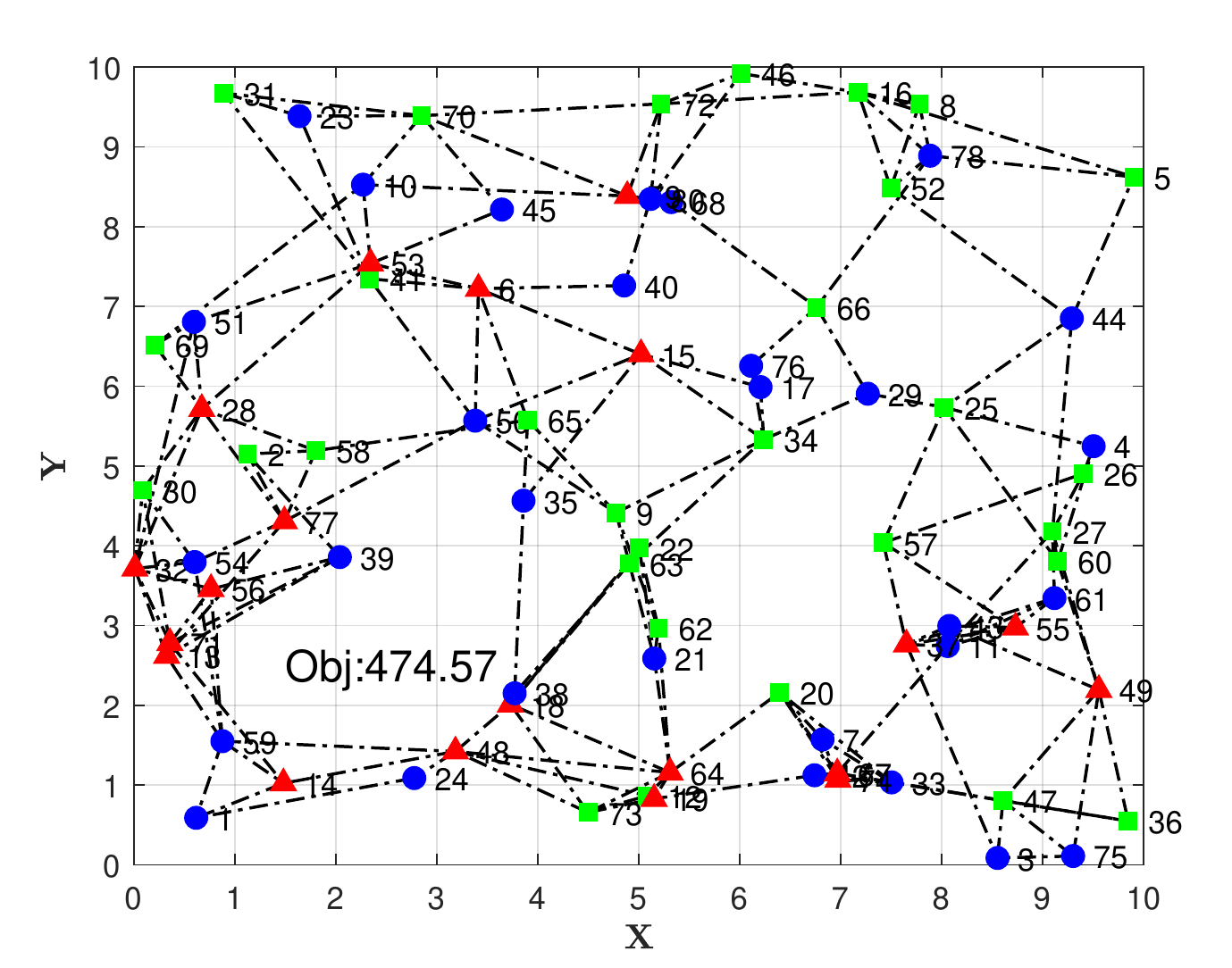}}
\subfigure[JayaX]{\includegraphics[width=1.7in]{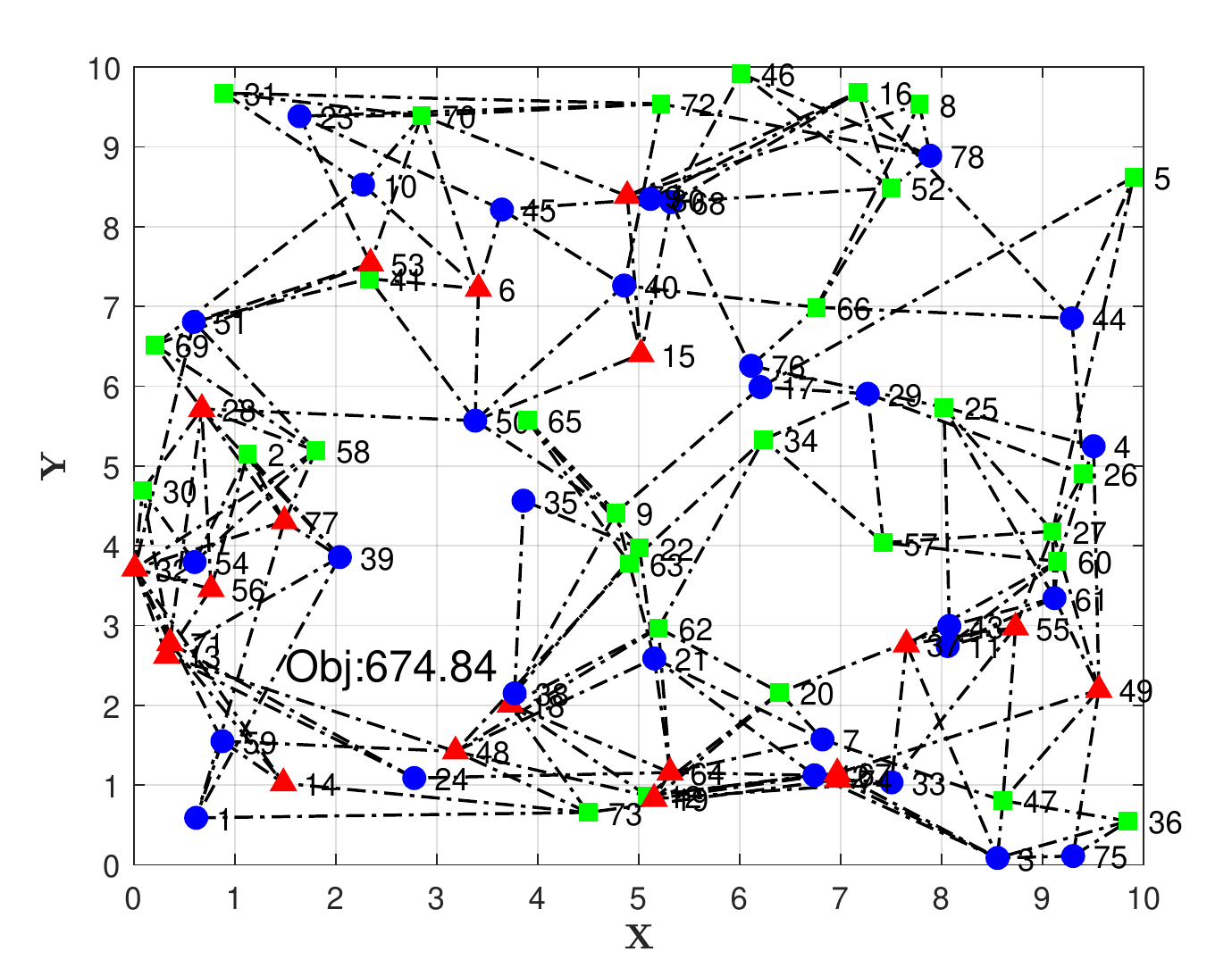}}
\subfigure[BPSO]{\includegraphics[width=1.7in]{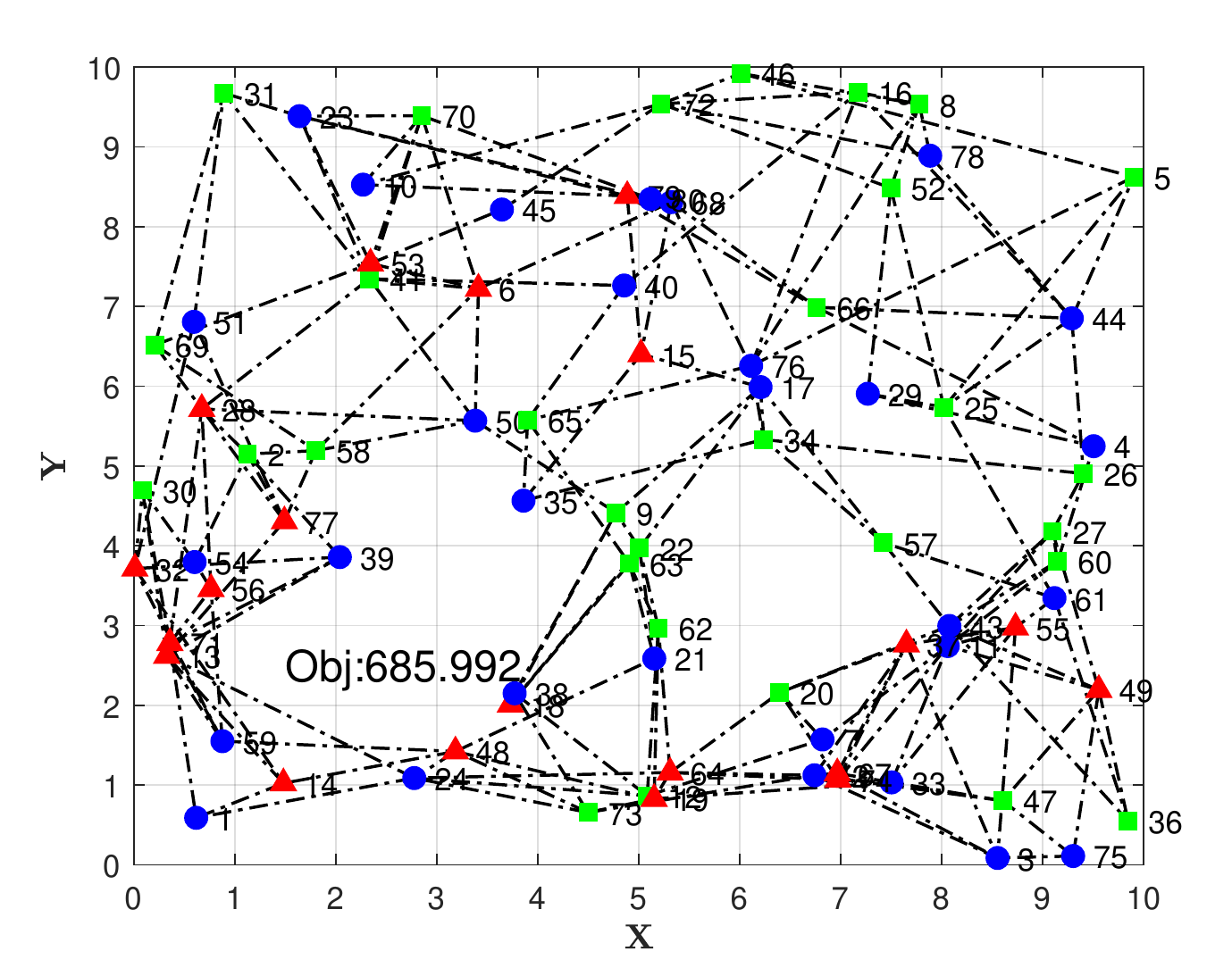}}
\subfigure[LBMDE]{\includegraphics[width=1.7in]{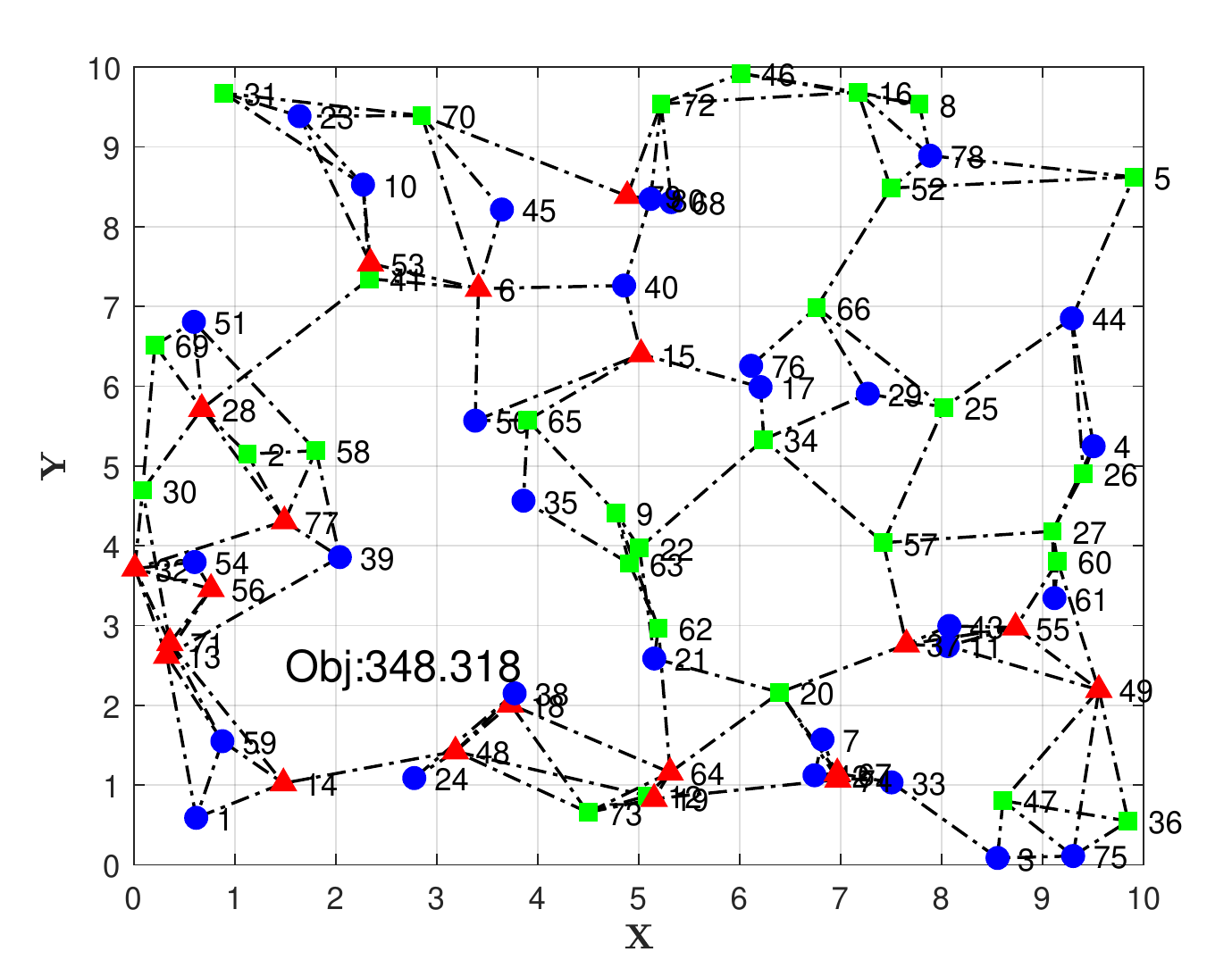}}
\subfigure{\includegraphics[width=2.3in]{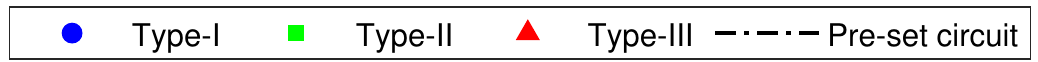}}

\caption{Optimal network structure for MNSDP-LIB-20-4 (the first six figures) and MNSDP-LIB-80-2 (the last six figures) test instances obtained by all compared algorithms.}
\label{fig_resulton20nodes}
\end{figure*}

\fref{fig_resulton20nodes} shows the optimal network structures obtained by all compared algorithms on MNSDP-LIB-20-4 and MNSDP-LIB-80-2. It's worth mentioning that, the results that are closest to the average objective values are chosen to present. As we can see, for problems with 20 nodes, almost all algorithms can find solutions with acceptable quality. However, for MNSDP-LIB-80-2, only LBMDE and CPLEX can get high-quality results. Due to the page limitation, other optimal network structures are not listed in this paper. Readers can find more information online.

\subsubsection{Computational complexity}
\begin{figure}[tbph]
	\begin{center}
		\includegraphics[width=3.5in]{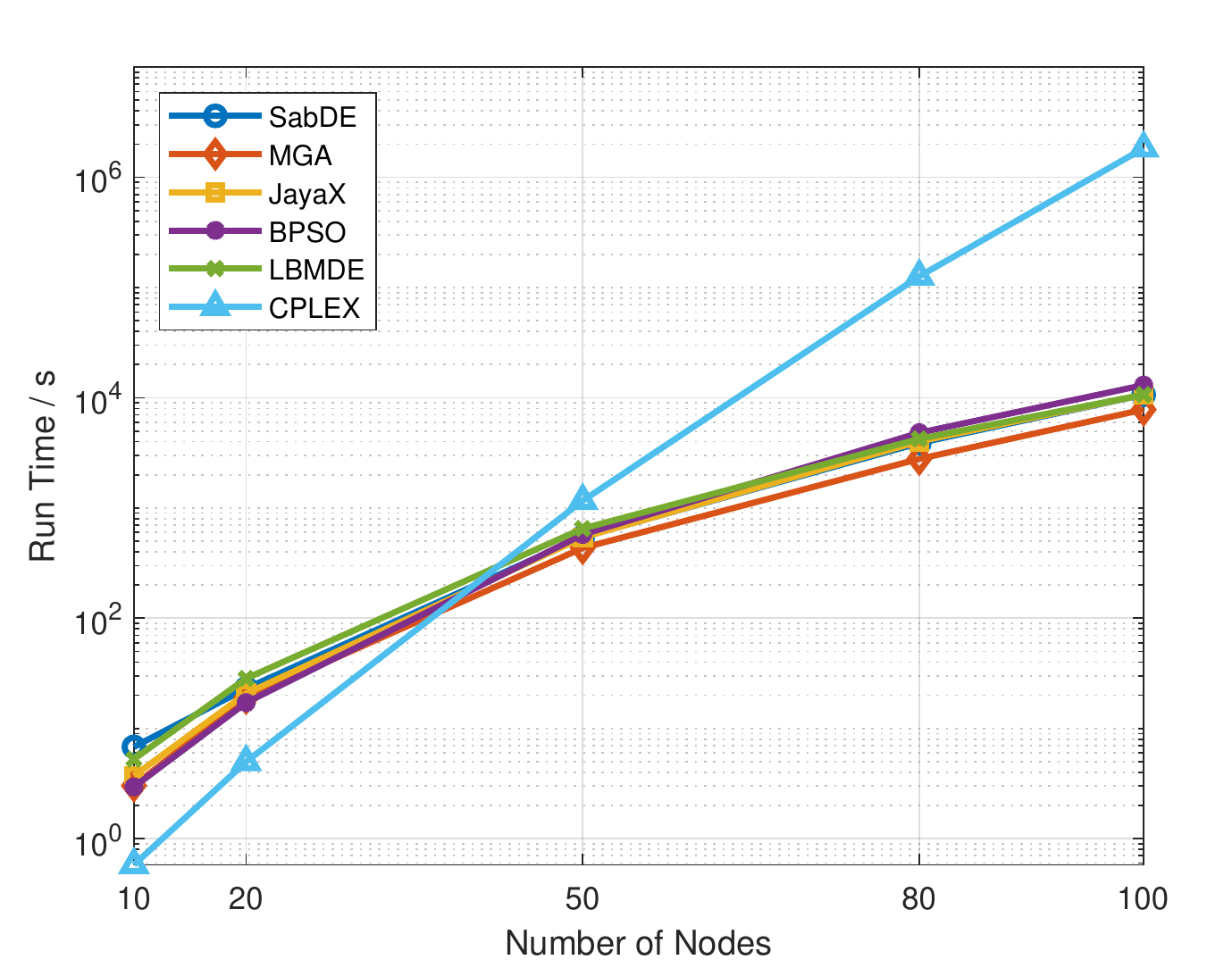}
		\caption{Average run time of all algorithms on different problems, where logarithmic axis is adopted in y-axis.}
		\label{fig_runtime}
	\end{center}
\end{figure}

In this part, we make an experimental computational complexity comparison between EAs and the traditional MIP method. To be specific, the average run time over 30 independent runs is collected for EAs, while the run time for MIP is presented by a single algorithm run. \fref{fig_runtime} shows the average running time of all algorithms on problems with different numbers of nodes. As we can observe, in terms of computational complexity, there is no significant difference between the existing EAs. That is, the run time of these EAs are highly related to the function evaluation. Specifically, BPSO is the fastest algorithm for small-scale problems while MGA obtains the best result for large-scale problems in terms of run time.

As we can see in \fref{fig_runtime}, for MIP methods like CPLEX, there is an exponential growth as the number of nodes increases (for problems with 80 nodes and 100 nodes, the gap is set to 3\% and 5\% respectively). For problems with 10 and 20 nodes, CPLEX shows its advantage in terms of run time. However, as the number of nodes increases, the growth of run time for EAs is slow and controllable. For large-scale problems, the run time is unbearable for MIP methods. In this situation, EAs become a feasible and effective tool. To sum up, the running time of all EAs is highly related to the number of nodes. CPLEX is competitive in dealing with small-scale problems while EAs show an overwhelming advantage when solving large-scale problems.

\subsubsection{Analysis on MNSDP-LIB}
In \sref{sec_benchmark}, we discussed the construction method of MNSDP-LIB in detail. To control the solving difficulty of test instances, parameter $r$ is introduced, which indicates the ratio of energy generation and consumption.

\begin{figure*}[!htb]
\setcounter{subfigure}{0}
\centering
\subfigure[MNSDP-LIB-10-1]{\includegraphics[width=1.3in]{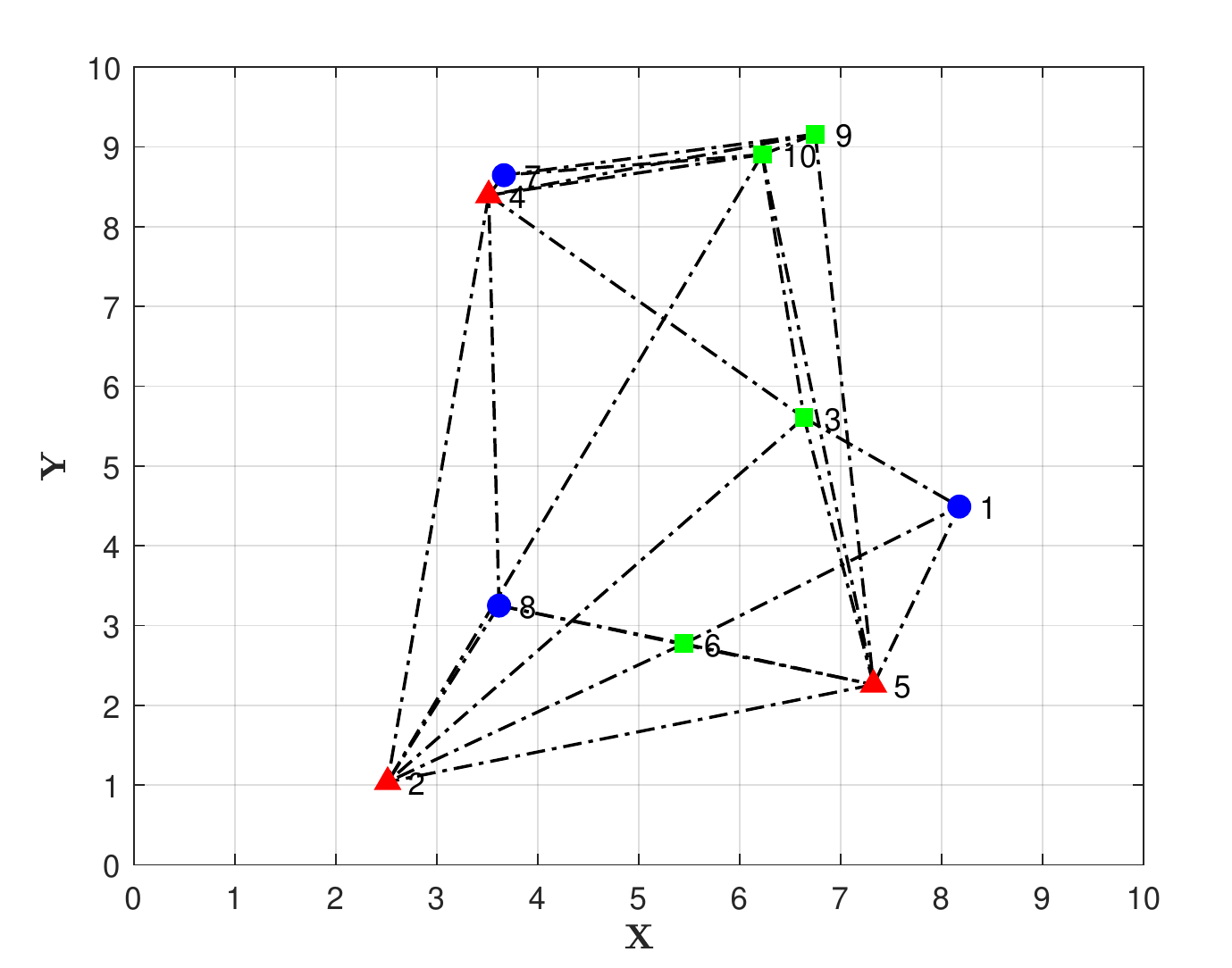}}
\subfigure[MNSDP-LIB-10-2]{\includegraphics[width=1.3in]{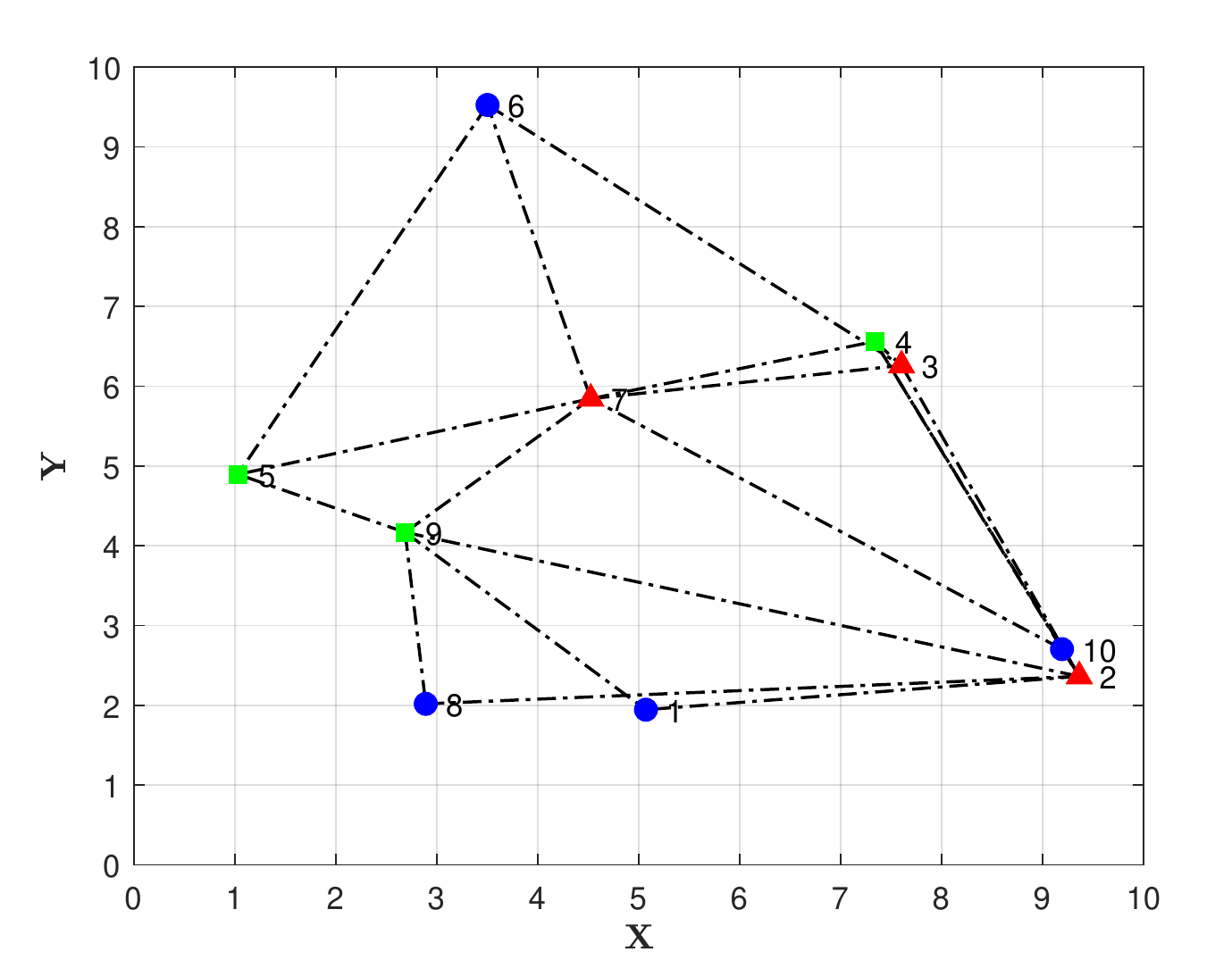}}
\subfigure[MNSDP-LIB-10-3]{\includegraphics[width=1.3in]{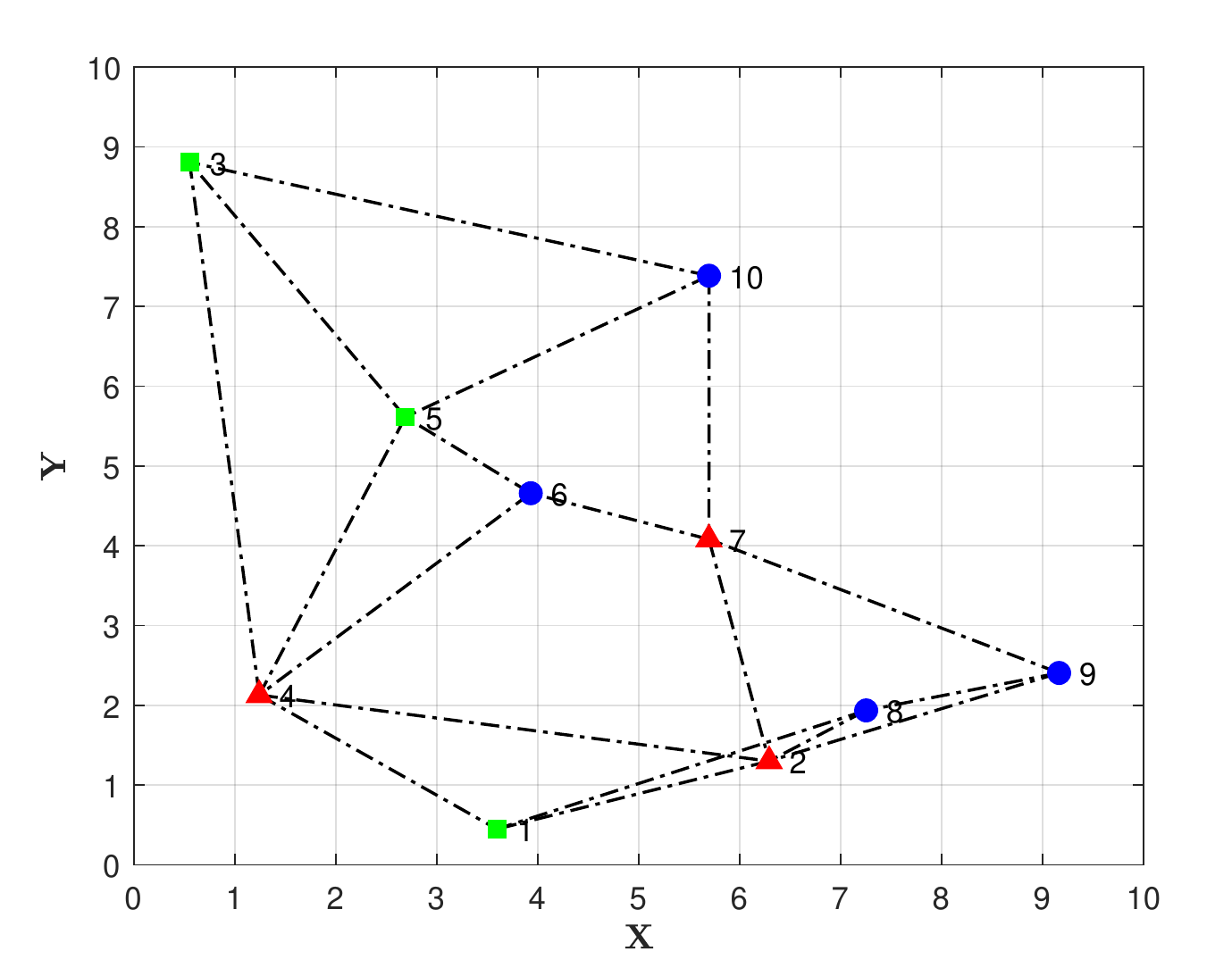}}
\subfigure[MNSDP-LIB-10-4]{\includegraphics[width=1.3in]{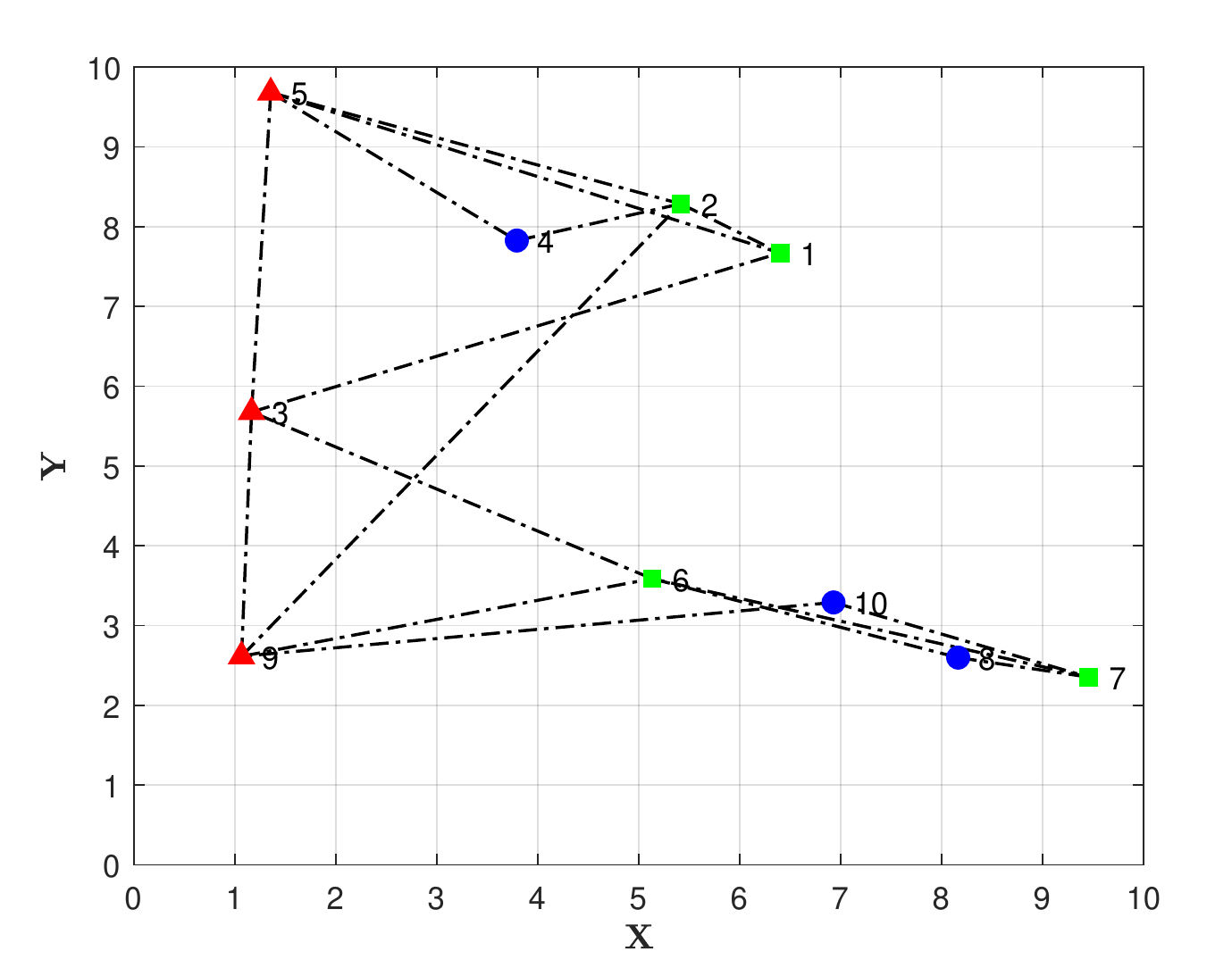}}
\subfigure[MNSDP-LIB-10-5]{\includegraphics[width=1.3in]{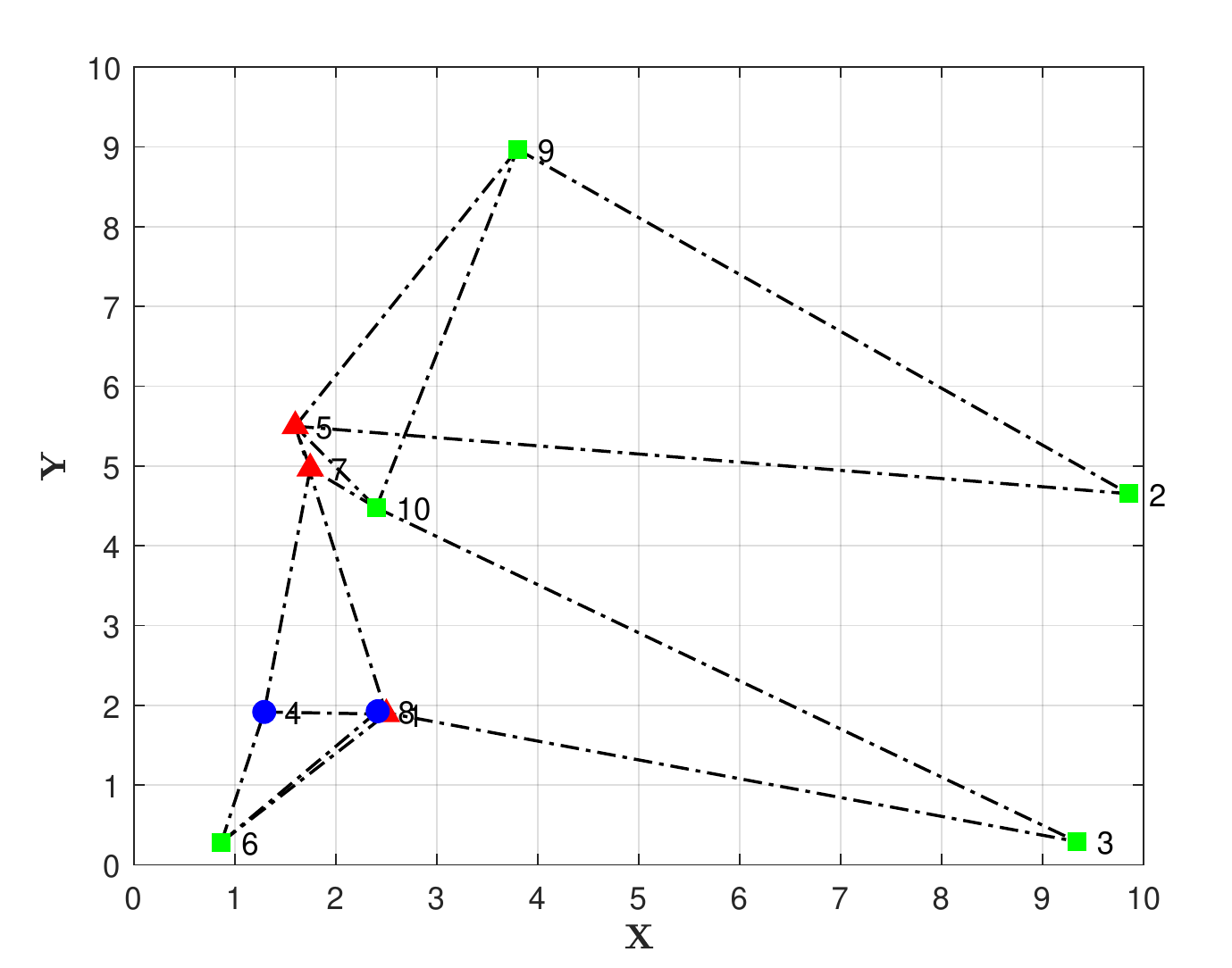}}

\subfigure[MNSDP-LIB-20-1]{\includegraphics[width=1.3in]{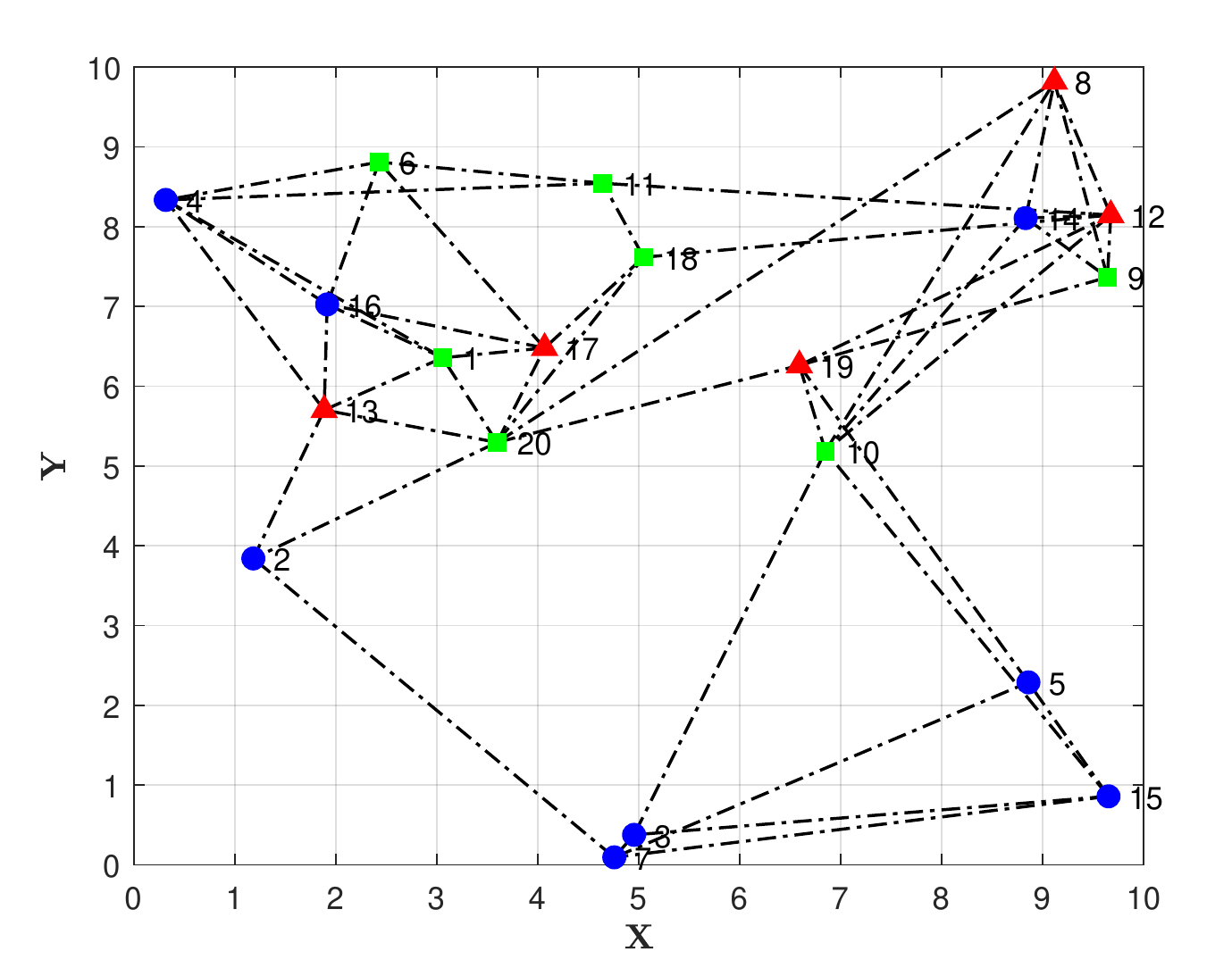}}
\subfigure[MNSDP-LIB-20-2]{\includegraphics[width=1.3in]{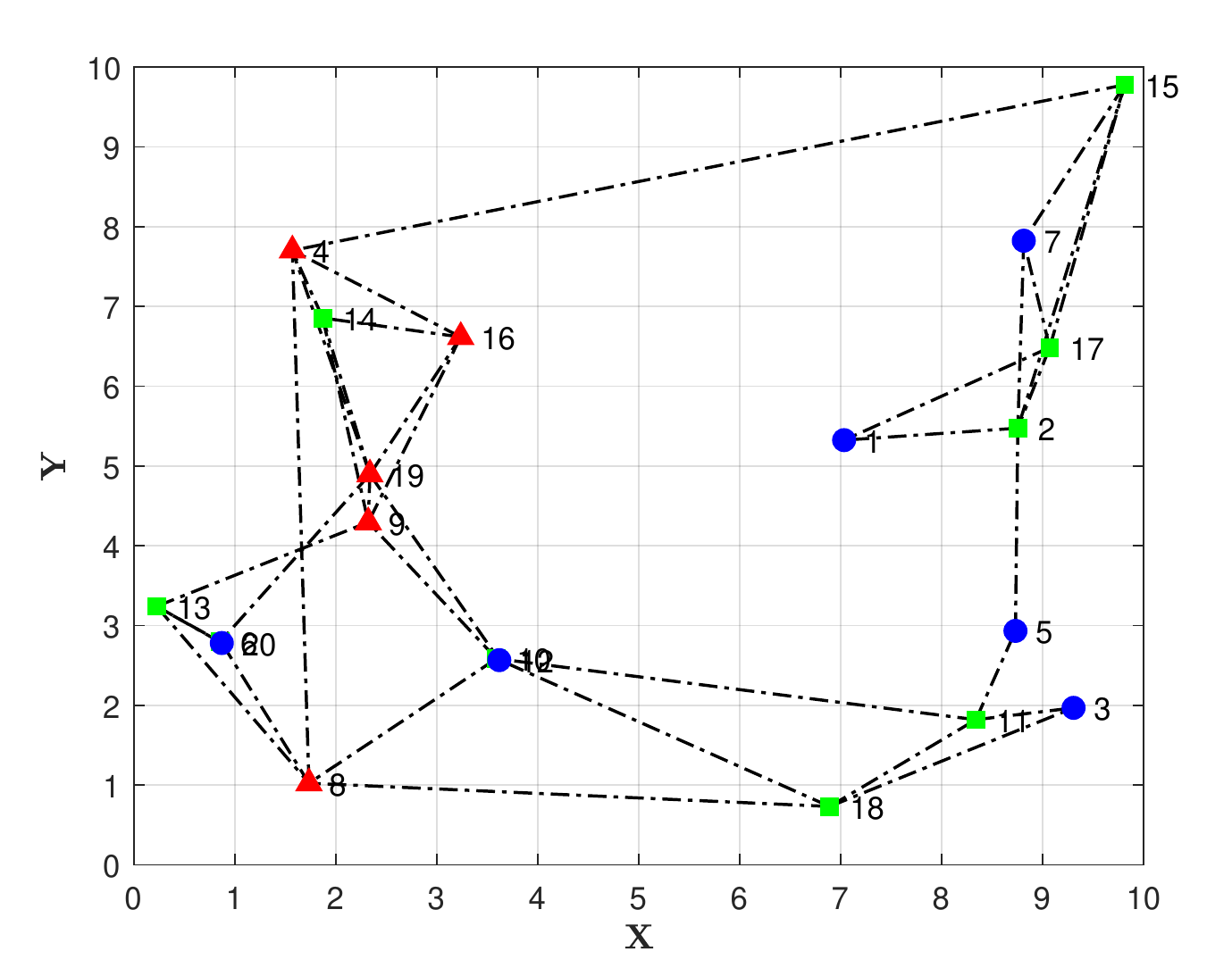}}
\subfigure[MNSDP-LIB-20-3]{\includegraphics[width=1.3in]{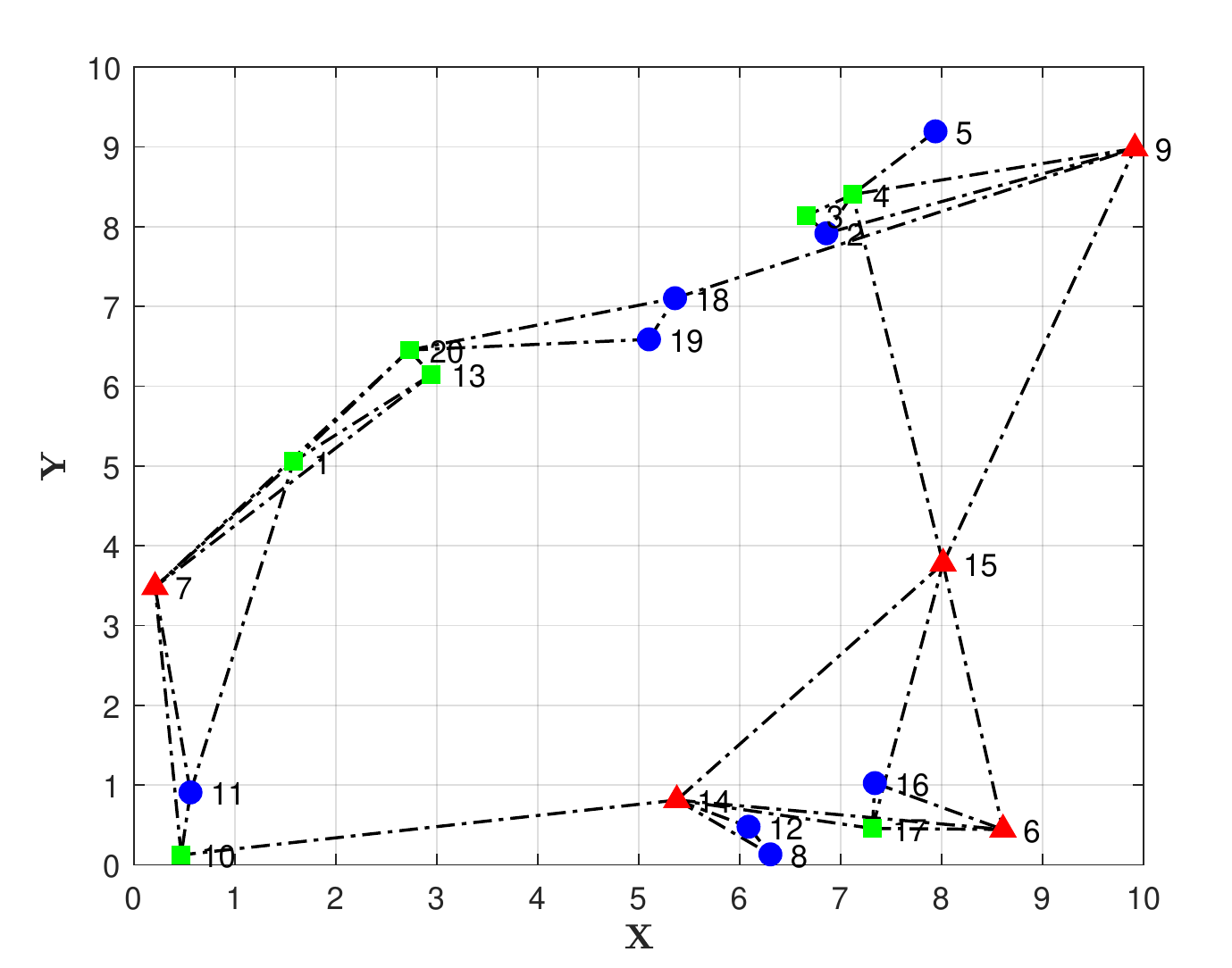}}
\subfigure[MNSDP-LIB-20-4]{\includegraphics[width=1.3in]{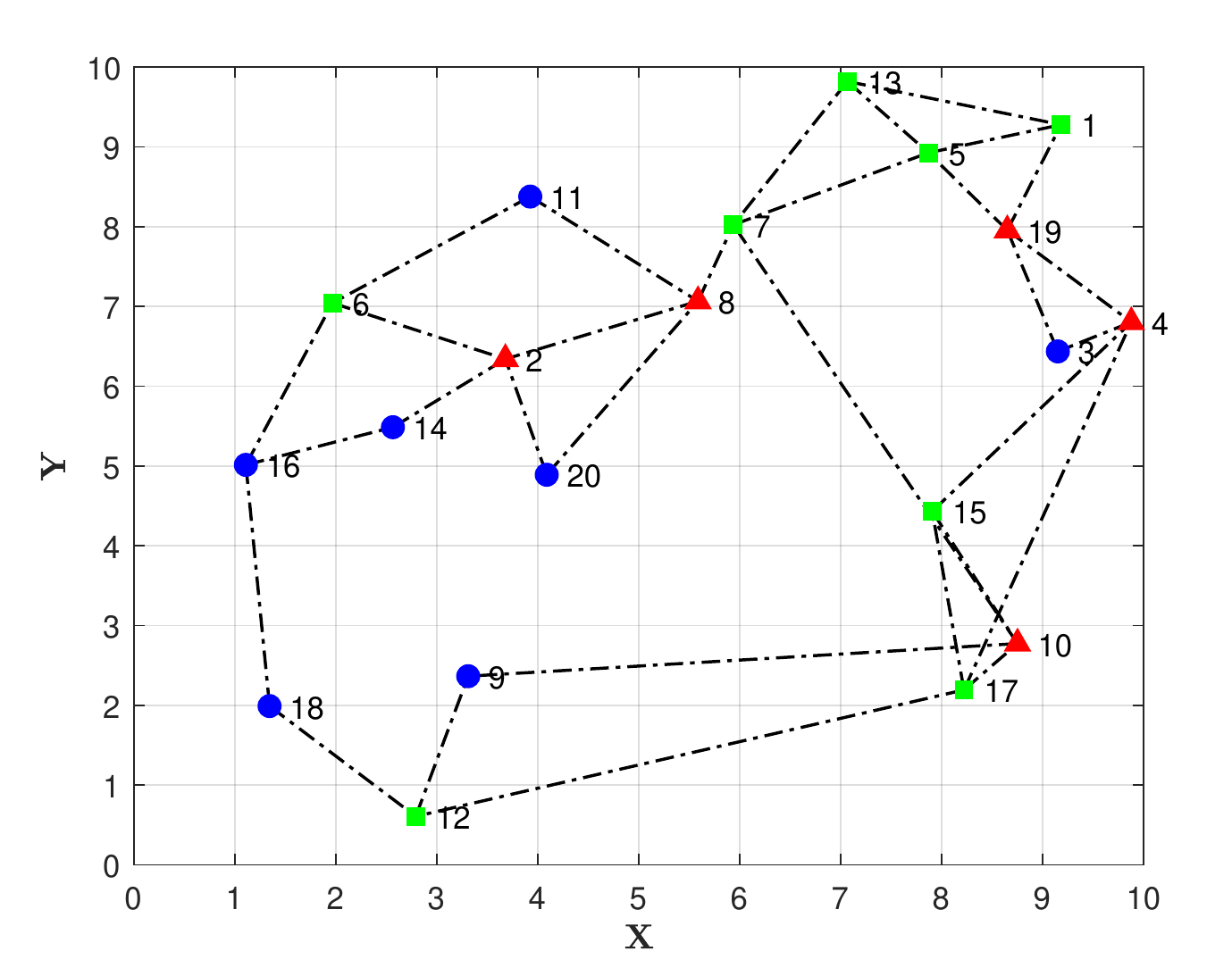}}
\subfigure[MNSDP-LIB-20-5]{\includegraphics[width=1.3in]{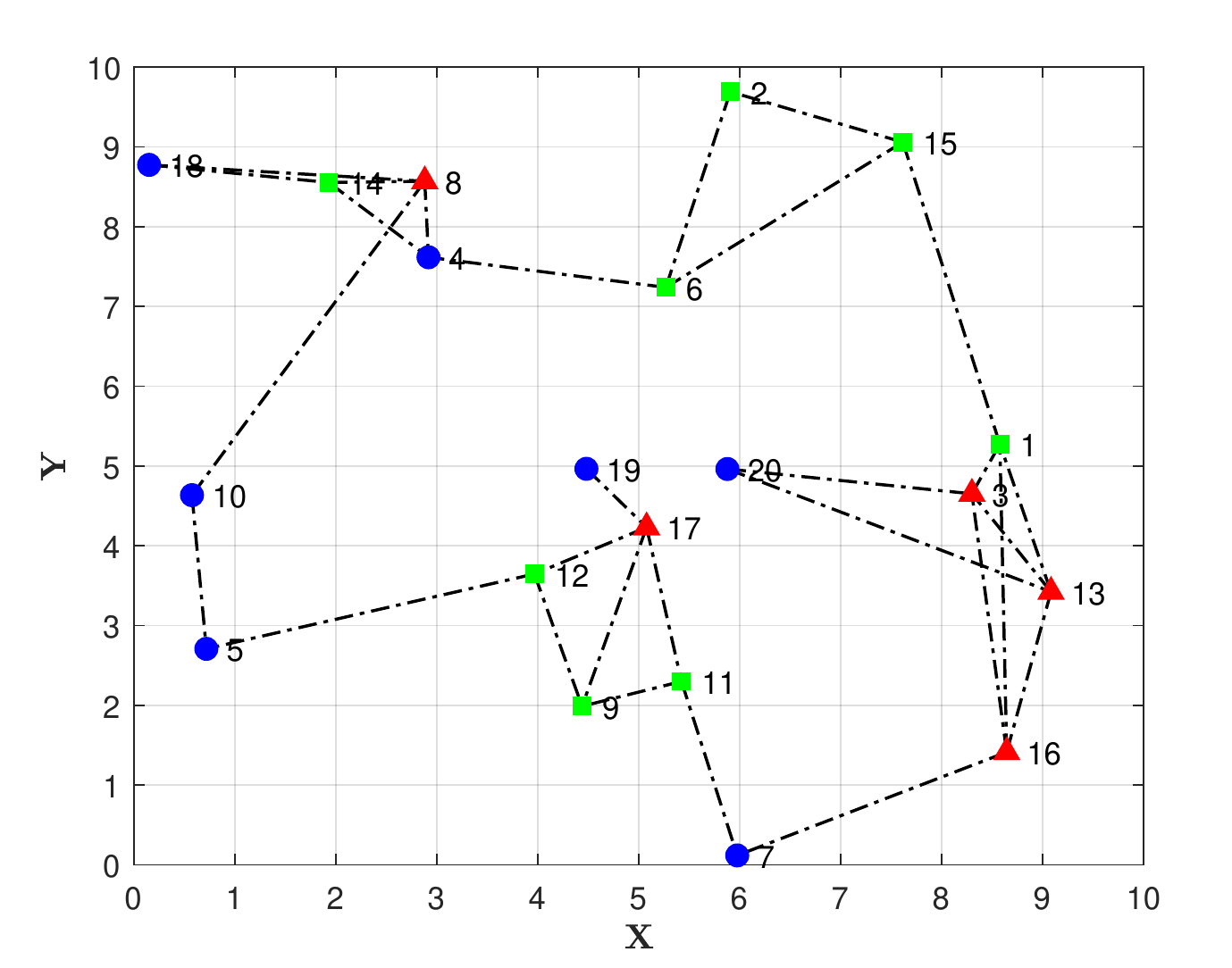}}

\subfigure[MNSDP-LIB-50-1]{\includegraphics[width=1.3in]{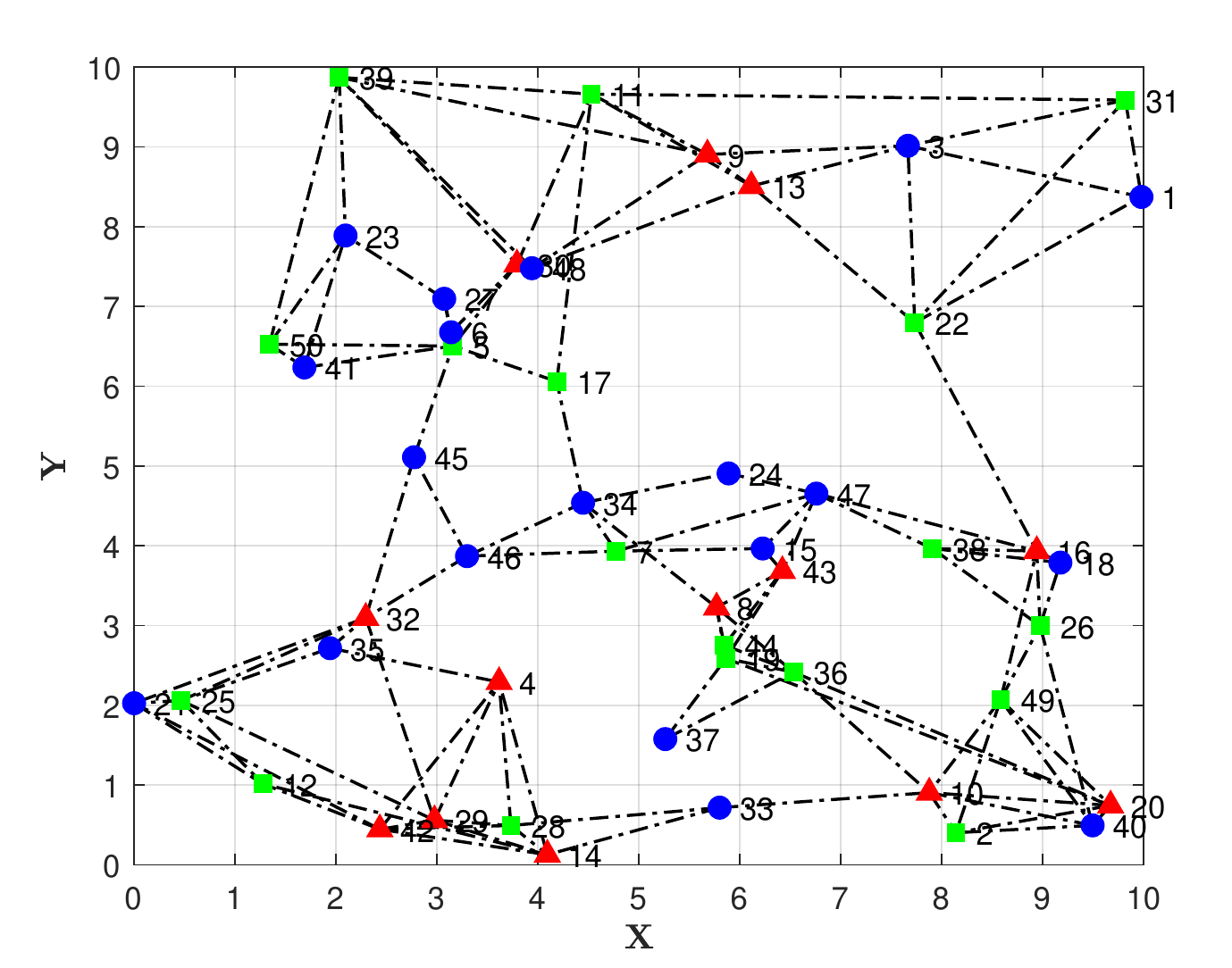}}
\subfigure[MNSDP-LIB-50-2]{\includegraphics[width=1.3in]{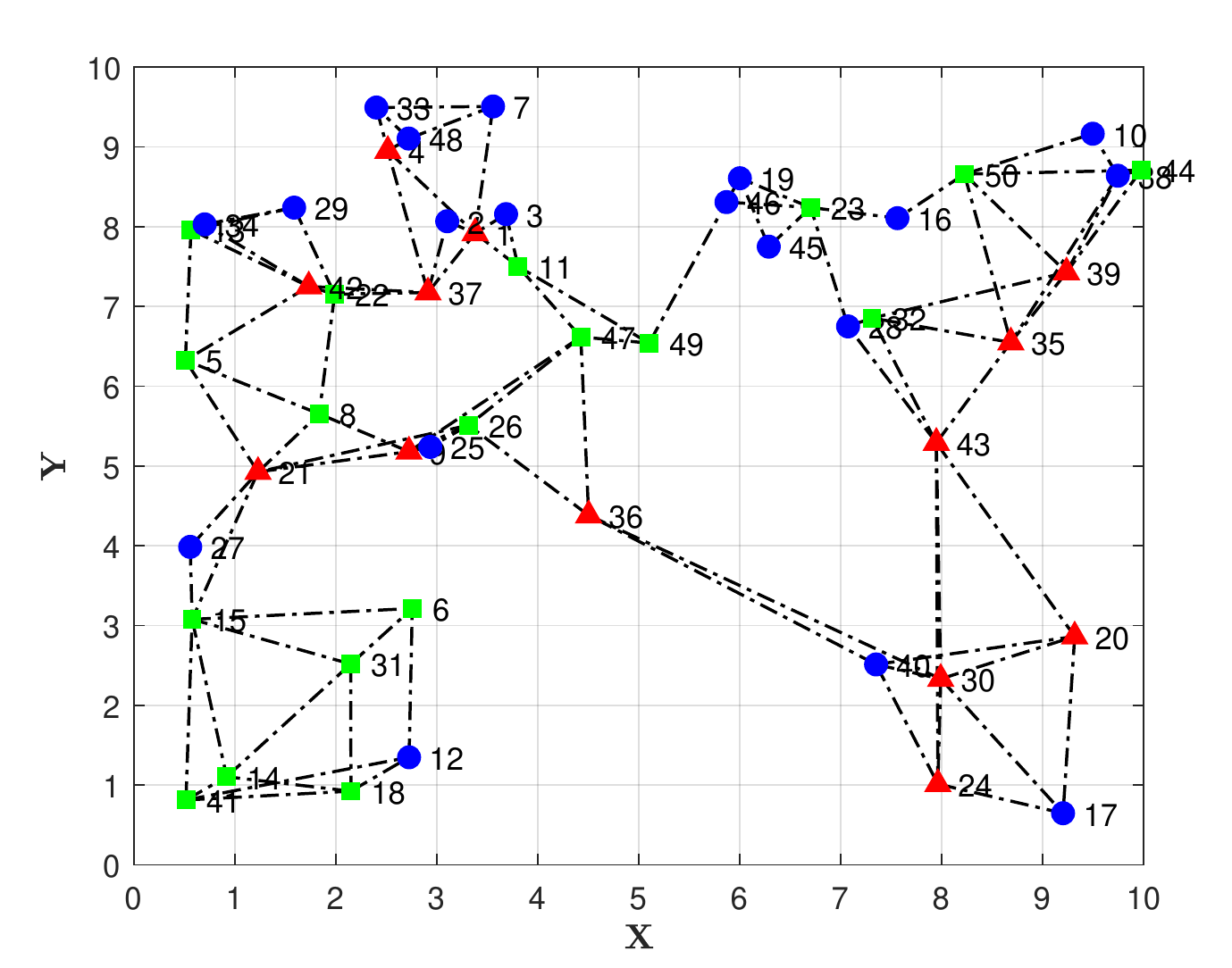}}
\subfigure[MNSDP-LIB-50-3]{\includegraphics[width=1.3in]{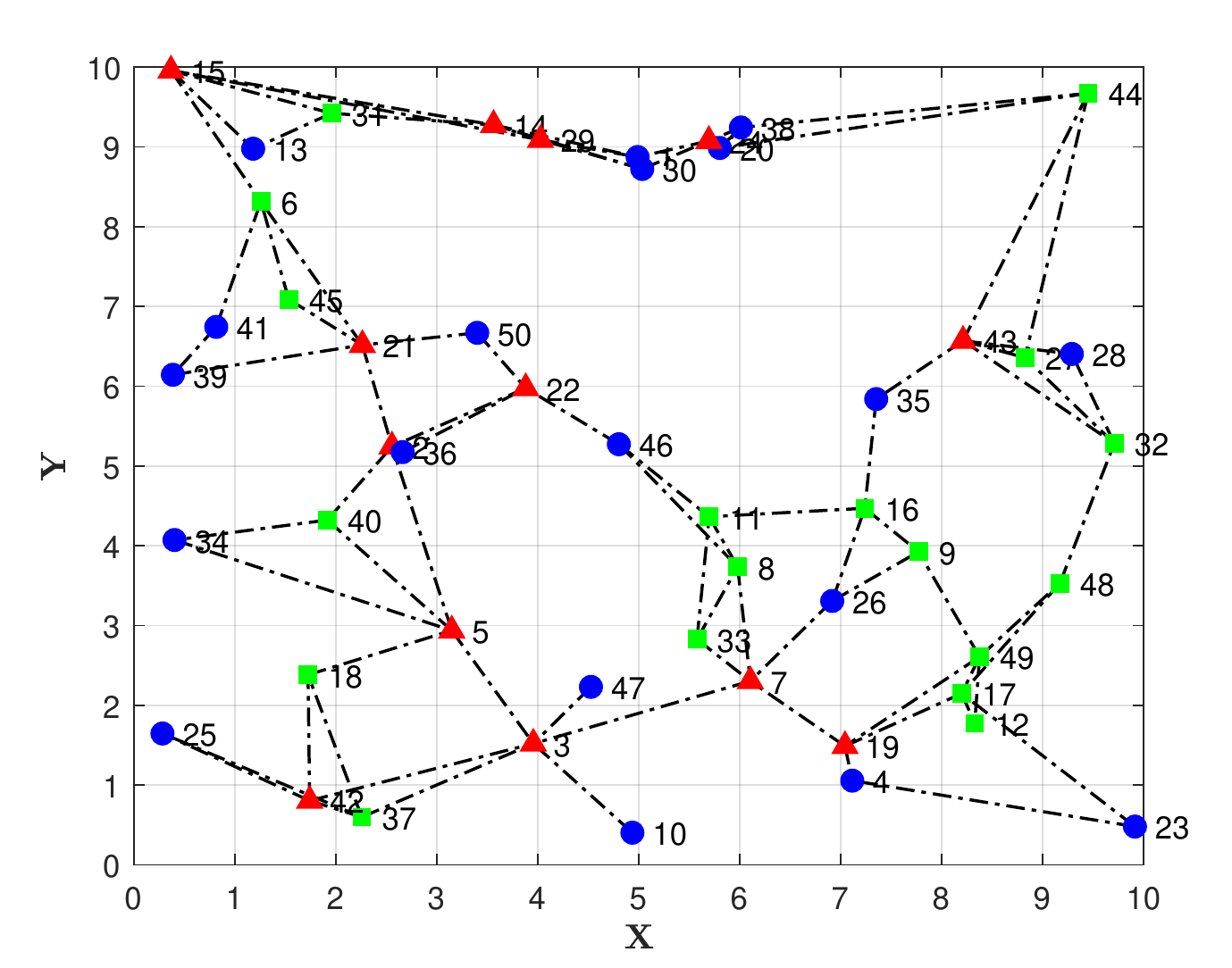}}
\subfigure[MNSDP-LIB-50-4]{\includegraphics[width=1.3in]{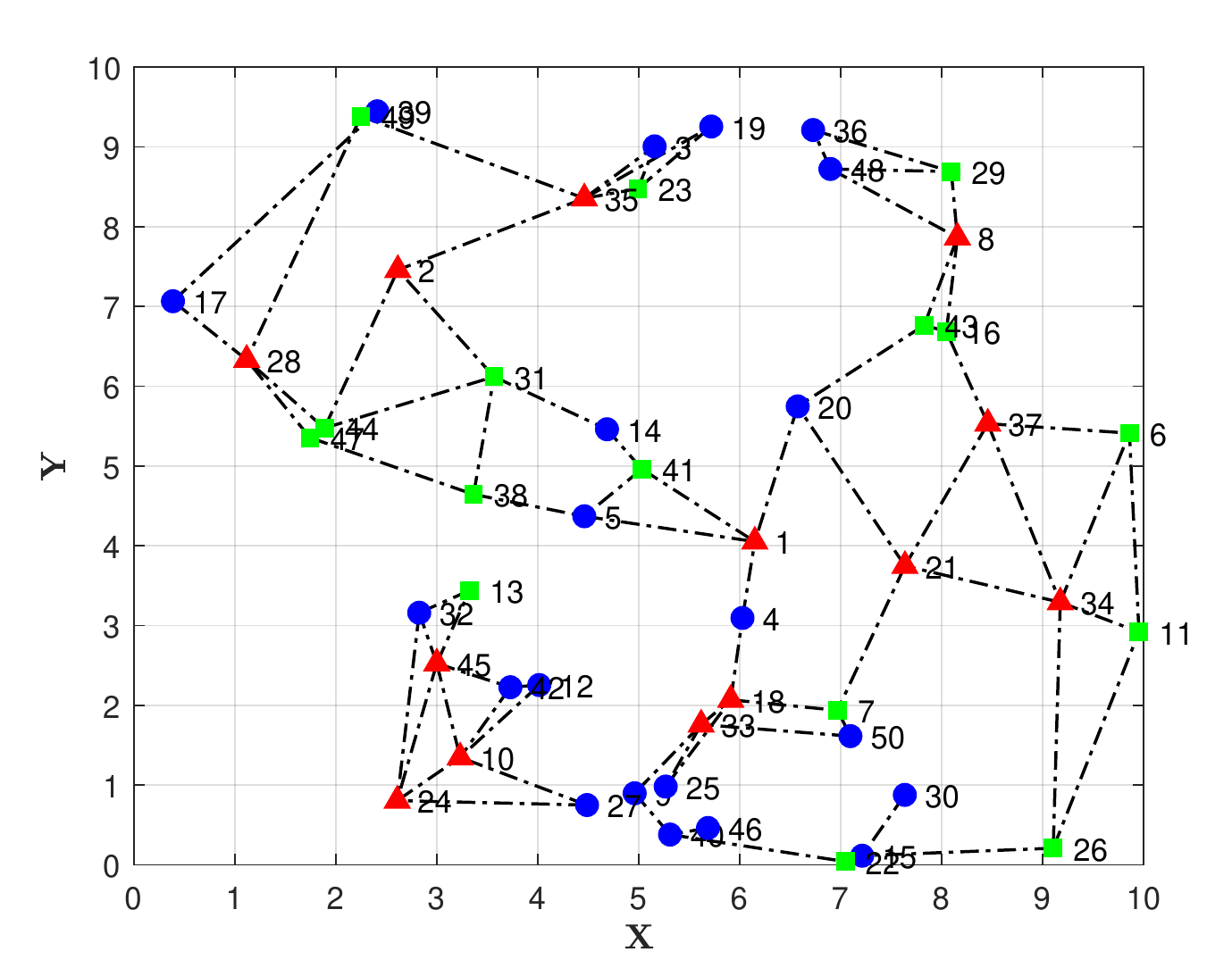}}
\subfigure[MNSDP-LIB-50-5]{\includegraphics[width=1.3in]{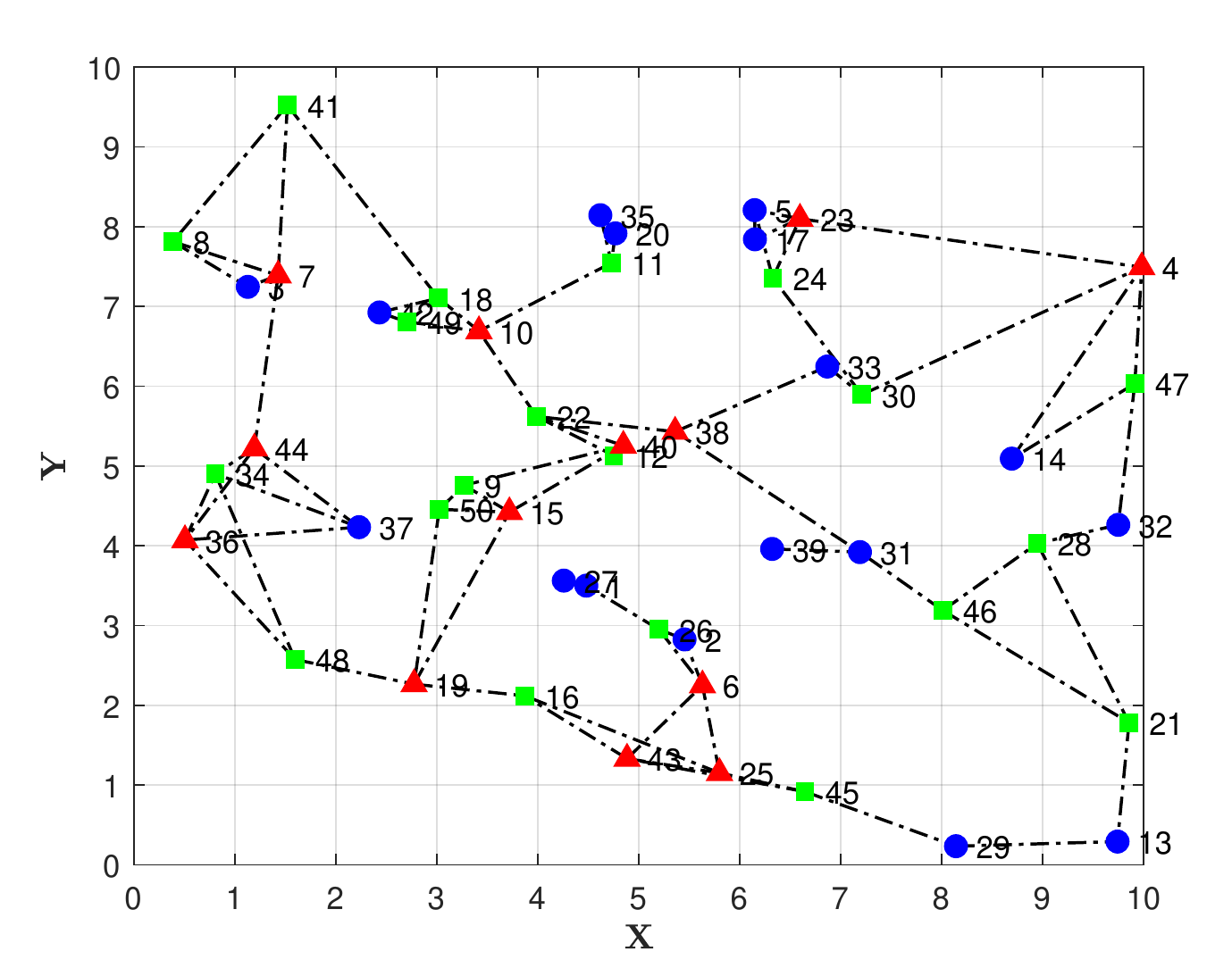}}
\subfigure{\includegraphics[width=2.3in]{legend.pdf}}
\caption{Optimal network structure for MNSDP-LIB-10, MNSDP-LIB-20 and MNSDP-LIB-50 test instances with different $r$ obtained by CPLEX.}
\label{fig_cplex_node50}
\end{figure*}

\fref{fig_cplex_node50} presents the final optimal power network structure for MNSDP-LIB-10, MNSDP-LIB-20 and MNSDP-LIB-50 test instances with different $r$. As we can see, with the increase of $r$, the optimal network structure becomes more and more simple. To be specific, the average number of neighbors for MNSDP-LIB-50-1 to MNSDP-LIB-50-5 are 4.56, 3.84, 3.32, 3 and 2.96 respectively. It means that under the precondition of system stability, nodes need to connect to fewer other nodes.

\subsection{Effect of the novel DE operator}

To further analyze the effect of the proposed novel DE operator, in this part, several variants of LBMDE (LBMbDE, LBMbPSO, LBMbGA and LBPJayaX) are proposed. Specifically, LBMbDE is developed based on LBMDE and replaces the novel DE operator with the basic DE operator according to \eref{equ_de}. LBMbGA adopts the simulated binary crossover and polynomial mutation operators. LBMbPSO and LBPJayaX are developed with binary PSO operator and JayaX operator. It's worth mentioning that, all other parameters are set according to LBMDE.

\begin{table*}
\centering
\caption{Average and standard variance results of the compared algorithms on MNSDP-LIB test suite, where the best mean for each test instance is highlighted.}
\begin{tabular}{cccccc}
\hline
\textbf{Problems} & \textbf{LBMbDE} & \textbf{LBMbPSO} & \textbf{LBMbGA} & \textbf{LBPJayaX} & \textbf{LBMDE} \bigstrut\\
\hline
\textbf{MNSDP-10-1} & 183.14(1.02) & 193.32(7.50) & 188.73(1.20) & 185.76(9.53) & \textbf{181.53(1.00)} \bigstrut[t]\\
\textbf{MNSDP-10-2} & 151.05(2.40) & 168.73(9.36) & \textbf{141.56(2.06)} & 144.11(6.70) & 144.29(2.38) \\
\textbf{MNSDP-10-3} & 123.64(0.86) & 146.67(8.25) & 124.97(1.03) & 133.06(5.19) & \textbf{122.65(0.44)} \\
\textbf{MNSDP-10-4} & 126.70(2.55) & 126.43(7.41) & 125.63(1.96) & 129.26(4.91) & \textbf{124.65(2.48)} \\
\textbf{MNSDP-10-5} & 114.57(0.21) & 119.06(7.02) & 118.05(1.83) & 113.00(4.99) & \textbf{113.48(0.20)} \\
\textbf{MNSDP-20-1} & 274.24(4.44) & 335.49(18.56) & 280.41(4.61) & 300.08(24.97) & \textbf{260.44(2.30)} \\
\textbf{MNSDP-20-2} & 202.51(2.78) & 238.84(15.29) & 202.50(1.85) & 252.28(22.18) & \textbf{190.47(2.77)} \\
\textbf{MNSDP-20-3} & 167.15(1.01) & 244.37(17.64) & 169.60(0.17) & 216.59(22.24) & \textbf{161.95(0.98)} \\
\textbf{MNSDP-20-4} & 148.57(1.47) & 179.60(10.89) & 154.18(1.31) & 154.24(12.42) & \textbf{144.36(1.44)} \\
\textbf{MNSDP-20-5} & 133.52(0.54) & 181.41(14.11) & 137.68(0.78) & 164.78(14.32) & \textbf{130.92(0.53)} \\
\textbf{MNSDP-50-1} & 435.83(2.44) & 586.62(143.99) & 435.70(4.28) & 545.33(36.43) & \textbf{387.39(2.39)} \\
\textbf{MNSDP-50-2} & 316.83(0.73) & 452.15(20.28) & 317.70(3.44) & 412.36(20.91) & \textbf{266.29(0.72)} \\
\textbf{MNSDP-50-3} & 303.16(1.27) & 402.61(188.69) & 291.08(3.76) & 371.20(19.06) & \textbf{250.41(1.26)} \\
\textbf{MNSDP-50-4} & 249.89(0.68) & 635.46(71.21) & 252.22(4.35) & 347.32(19.33) & \textbf{203.71(0.66)} \\
\textbf{MNSDP-50-5} & 238.29(0.59) & 621.74(59.77) & 234.86(4.56) & 312.05(17.86) & \textbf{190.96(0.57)} \\
\textbf{MNSDP-80-1} & 545.94(2.89) & 3232.15(204.93) & 597.65(5.90) & 785.26(37.78) & \textbf{436.33(2.80)} \\
\textbf{MNSDP-80-2} & 465.29(3.16) & 664.30(21.55) & 476.02(6.31) & 609.62(27.92) & \textbf{348.42(3.15)} \\
\textbf{MNSDP-80-3} & 396.70(2.73) & 2381.99(426.26) & 428.65(5.78) & 561.11(31.69) & \textbf{301.66(2.70)} \\
\textbf{MNSDP-80-4} & 381.40(1.79) & 2511.02(174.32) & 438.03(6.25) & 564.96(19.53) & \textbf{267.46(1.70)} \\
\textbf{MNSDP-80-5} & 337.91(2.50) & 2416.09(171.19) & 382.72(5.87) & 449.57(28.03) & \textbf{239.71(2.46)} \bigstrut[b]\\
\hline
\end{tabular}%

\label{tab_result2}
\end{table*}

\tref{tab_result2} lists the results obtained by all LBMDE variants algorithms on MNSDP-LIB test instances. As we can see, among all compared algorithms, LBMDE receives the best performance. Specifically, LBMbDE and LBMbGA get high-quality solutions. Compared to SabDE and MGA, the variants have some advantages in dealing with large-scale problems. In \sref{sec_sel}, we proposed a novel environmental selection strategy to better handle the constraints. As a result, algorithms can receive better results. For LBMbPSO, the final obtained results are relatively weak. As we discussed in the previous section, the updating process of PSO is not suitable for solving large-scale sparse problems, which will quickly converge and the diversity of solutions in the decision space is poor. Therefore, a novel strategy for implementing PSO to solve such a problem is needed.

\begin{figure}[!htb]
\setcounter{subfigure}{0}
\centering
\subfigure[MNSDP-LIB-20-3]{\includegraphics[width=1.7in]{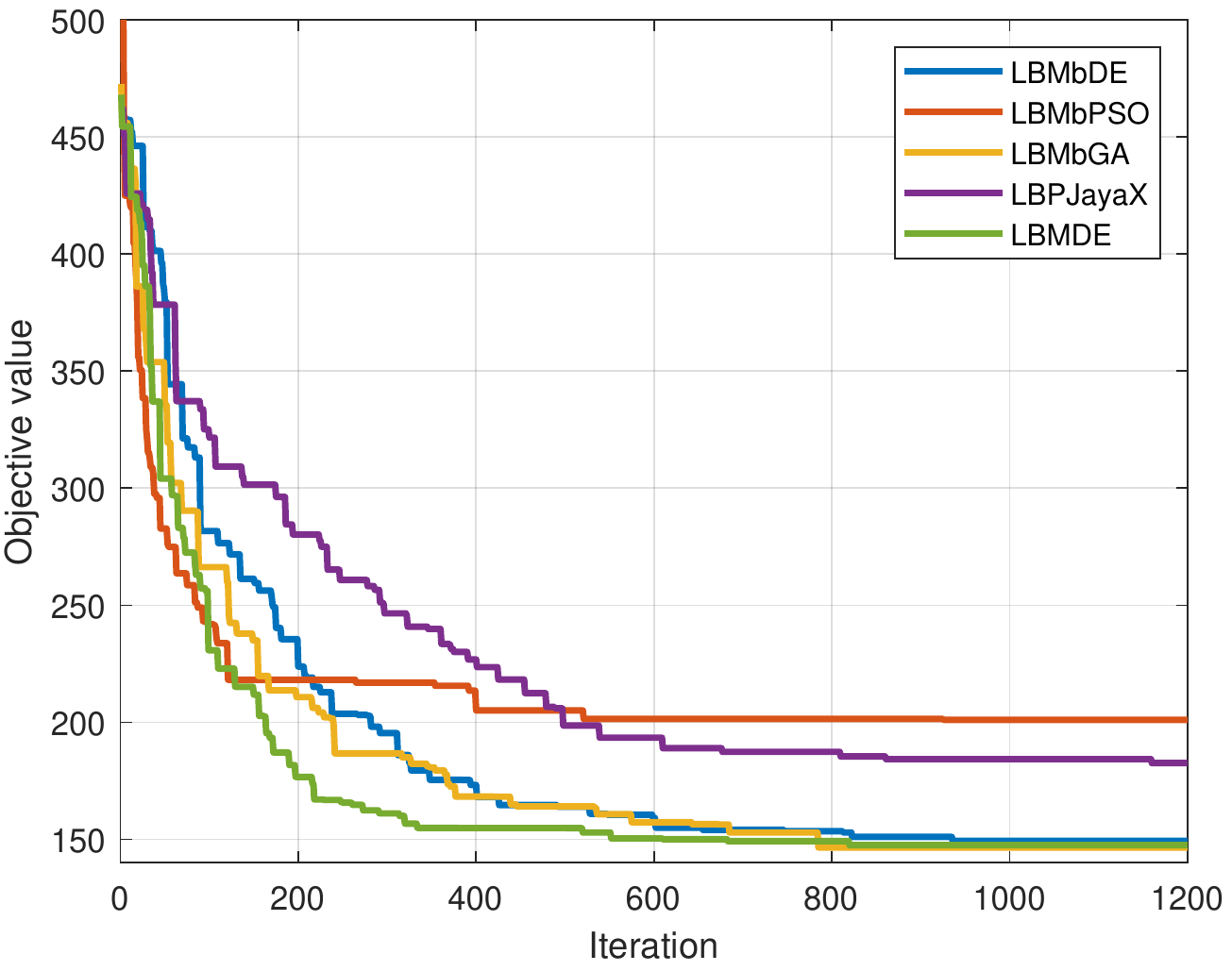}}
\subfigure[MNSDP-LIB-50-3]{\includegraphics[width=1.7in]{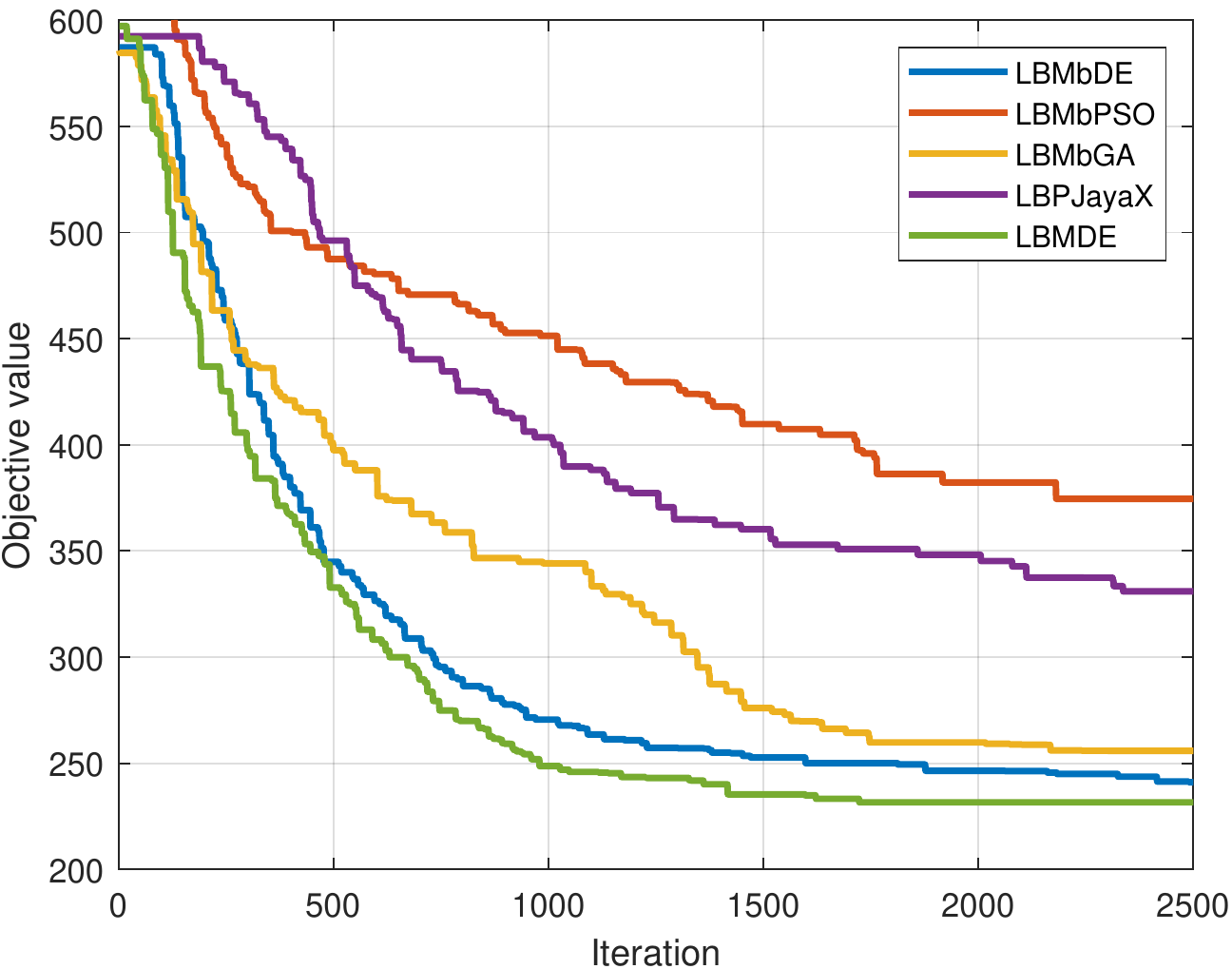}}

\caption{Convergence process of LBMDE variants on solving MNSDP-LIB with 20 nodes (left) and 50 nodes (right).}
\label{fig_conopera}
\end{figure}

In addition, \fref{fig_conopera} shows the convergence process of all compared algorithms for solving MNSDP-LIB-20-3 and MNSDP-LIB-50-3. As we can see, for the low-dimension problem, both LBMbDE, LBMbGA and LBMDE can obtain the optimal solution. To be specific, LBMDE has a faster convergence speed than LBMbDE and LBMbGA. For LBMbPSO, it can quickly evolve in the early stage of the searching process. However, it seems that LBMbPSO can easily get trapped into local optima in the early stage. For LBPJayaX, the convergence process is slow but stable. The updating process of LBPJayaX is at the level of individual decision variables. For large-scale problems that exist a high correlation between decision variables, the JayaX operator performs poorly.

\section{Conclusion}
\label{sec_conclusion}
Many real-world engineering problems can be categorized as binary matrix optimization problems (BMOPs). However, few studies focus on solving this kind of problem. In this study, the large-scale network structure design problem is analyzed, which is encoded with a binary matrix. For many EAs that are designed for large-scale optimization problems, dividing decision variables into several groups and optimizing them separately is the basic idea to accelerate the searching process. Therefore, figuring out the grouping method is important, which is usually based on the co-relation of decision variables. For large-scale BMOPs, decision variables in the same row/column are naturally categorized into one group. As a result, it's reasonable to regard them as a whole to perform the crossover and mutation operations.

Based on the above-mentioned idea, we proposed a binary-matrix-based DE operator for BMOPs and an improved feasible rules based environmental selection method, resulting in a novel constrained differential evolutionary algorithm (LBMDE). Then, a multi-microgrid network structure design problem is adopted to examine the performance of the proposed and other existing state-of-the-art EAs.

The proposed LBMDE shows competitive performance in solving large-scale multi-microgrid network structure design problems compared to the commercial solver and other representative EAs. In addition, such ideas proposed in this study can be easily extended to solve other large-scale BMOPs, which is also one of our future works.

\bibliographystyle{IEEEtran}
\bibliography{reference}

\begin{IEEEbiography}[{\includegraphics[width=1in,height=1.25in]{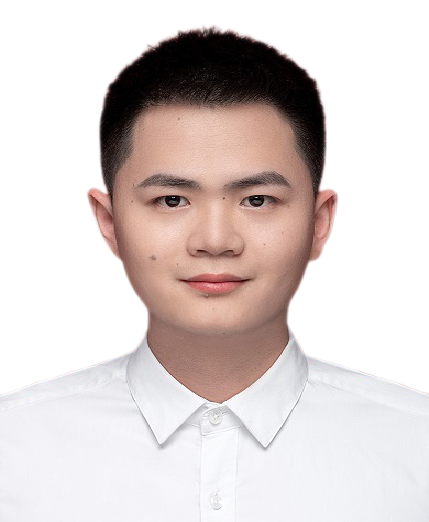}}]{Wenhua Li} received his B.S. and M.S. degrees in 2018 and 2020, respectively, from 	National University of Defense Technology (NUDT), Changsha, China. He is now a Ph.D. student in Management Science and Technology. His current research interests include multi-objective evolutionary algorithms, energy management in microgrids and artificial intelligence.
\end{IEEEbiography}
\vspace{-20 pt}

%
%

\begin{IEEEbiography}[{\includegraphics[width=1in,height=1.25in]{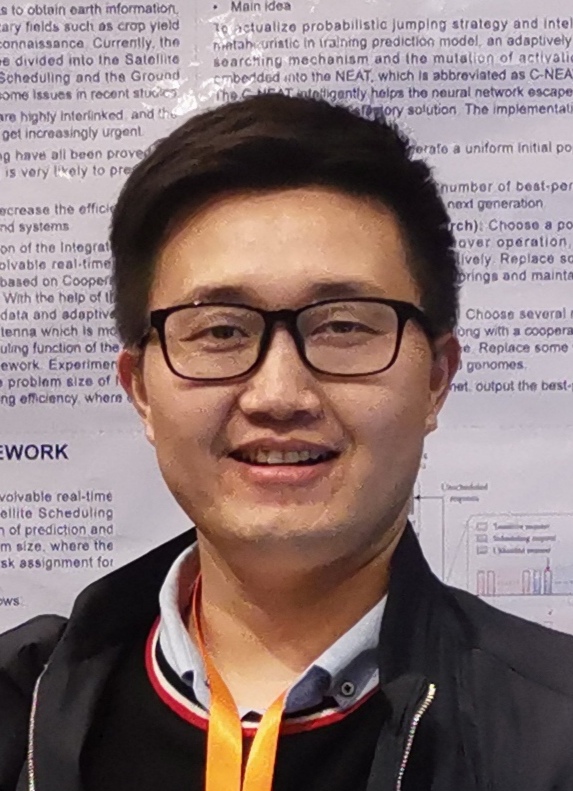}}]{Rui Wang} (Senior Member, IEEE) received his Bachelor degree from the National University of Defense Technology, P.R. China in 2008, and the Doctor degree from the University of Sheffield, U.K in 2013. Currently, he is an Associate professor with the National University of Defense Technology. His current research interest includes evolutionary computation, multi-objective optimization and the development of algorithms applicable in practice.
	
Dr. Wang received the Operational Research Society Ph.D. Prize at 2016, and the National Science Fund for Outstanding Young Scholars at 2021. He is also an Associate Editor of the Swarm and Evolutionary Computation, the IEEE Trans on Evolutionary Computation.
\end{IEEEbiography}


\end{document}